\documentclass[10pt,twocolumn,letterpaper]{article}

\usepackage{cvpr}
\usepackage{times}
\usepackage{epsfig}
\usepackage{graphicx}
\usepackage{amsmath}
\usepackage{amssymb}
\usepackage{subcaption}
\usepackage{scalefnt}
\usepackage{placeins}

\graphicspath{{figures/}}
\newcommand{\modelname}{AvatarMe}
\newcommand{\datasetname}{RealFace\textsc{DB}}
\newcommand\smaller[2][0.8]{{\scalefont{#1}#2}}
\newcommand{\threedmm}{\smaller{3}\textsc{dmm}}
\newcommand{\threed}{\smaller{3}\textsc{d}}
\newcommand{\twod}{\smaller{2}\textsc{d}}
\newcommand{\fourd}{\smaller{4}\textsc{d}}

\newcommand{\renderready}{render-ready}

\usepackage[pagebackref=true,breaklinks=true,letterpaper=true,colorlinks,bookmarks=false]{hyperref}

\cvprfinalcopy 

\def\cvprPaperID{5793} 

\ifcvprfinal\fi

\makeatletter
\@namedef{ver@everyshi.sty}{}
\makeatother
\usepackage{pgffor}
\begin{document}

\title{\modelname: Realistically Renderable 3D Facial Reconstruction ``in-the-wild''}

\author{Alexandros Lattas$^{1,2}$
       \hspace{0.8cm}
       Stylianos Moschoglou$^{1,2}$
       \hspace{0.8cm}
       Baris Gecer$^{1,2}$
       \hspace{0.8cm}
       Stylianos Ploumpis$^{1,2}$
       \\
       Vasileios Triantafyllou$^{2}$
       \hspace{0.8cm}
       Abhijeet Ghosh$^{1}$
       \hspace{0.8cm}
       Stefanos Zafeiriou$^{1,2}$
\and
$^1$Imperial College London, UK
\hspace{0.8cm}  
$^2$FaceSoft.io
\\
{\tt\scriptsize $^1$\{a.lattas,s.moschoglou,b.gecer,s.ploumpis,ghosh,s.zafeiriou\}@imperial.ac.uk}
\hspace{0.2cm}
{\tt\scriptsize $^2$v.triantafyllou@facesoft.io}
}

\twocolumn[{%
\renewcommand\twocolumn[1][]{#1}%
\maketitle
\begin{center}
    \centering
    \includegraphics[width=\textwidth]{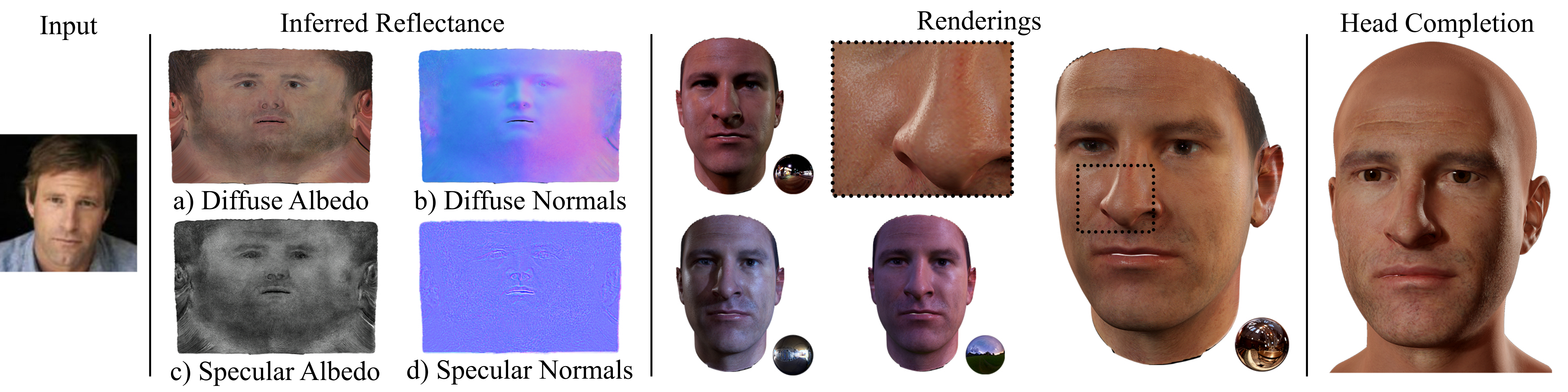}
    \captionof{figure}{
        From left to right:
        Input image;
        Predicted reflectance (diffuse albedo, diffuse normals,
        specular albedo and specular normals);
        Rendered reconstruction in different environments, with detailed reflections;
        Rendered result with head completion.
    }
    \label{fig:teaser}
\end{center}%
}]

\begin{abstract}
\vspace{-0.35cm}
Over the last years,
with the advent of Generative Adversarial Networks (\textsc{gan}s), 
many face analysis tasks have accomplished astounding performance,
with applications including, but not limited to, 
face generation and \threed{} face reconstruction from a single ``in-the-wild'' image. 
Nevertheless, to the best of our knowledge, 
there is no method which can produce high-resolution photorealistic \threed{} faces from ``in-the-wild'' images 
and this can be attributed to the: 
(a) scarcity of available data for training, 
and (b) lack of robust methodologies 
that can successfully be applied on very high-resolution data. 
In this paper, we introduce \modelname, 
the first method that is able to reconstruct photorealistic \threed{} faces
from a single ``in-the-wild'' image with an increasing level of detail.
To achieve this, 
we capture a large dataset of facial shape and reflectance 
and build on a state-of-the-art \threed{} texture and shape reconstruction method 
and successively refine its results, while generating 
the per-pixel diffuse and specular components 
that are required for realistic rendering. 
As we demonstrate in a series of qualitative and quantitative experiments, 
\modelname{} outperforms the existing arts by a significant margin 
and reconstructs authentic, 
\smaller{4}\textsc{k} by \smaller{6}\textsc{k}-resolution \threed{} faces from a single
low-resolution image that, 
for the first time, bridges the uncanny valley.
\end{abstract}


\section{Introduction}

The reconstruction of a \threed{} face geometry and texture 
is one of the most popular and well-studied fields 
in the intersection of computer vision, graphics and machine learning. 
Apart from its countless applications,
it demonstrates the power of recent developments
in scanning, learning and synthesizing \threed{} objects
\cite{blanz1999morphable, zhou2019dense}. 
Recently, mainly due to the advent of deep learning, tremendous progress has been made in the reconstruction of a smooth \threed{} face geometry, even from images captured in arbitrary recording conditions (also referred to as ``in-the-wild'') \cite{gecer2019synthesizing, gecer2019ganfit, sela2017unrestricted, tewari2018self, tran2019learning}. 
Nevertheless, even though the geometry can be inferred somewhat accurately, in order to render a reconstructed face in arbitrary virtual environments, much more information than a \threed{} smooth geometry is required, i.e., skin reflectance as well as high-frequency normals. 
In this paper, we propose a meticulously designed pipeline for the reconstruction of high-resolution \renderready{} faces from ``in-the-wild'' images captured in arbitrary poses, lighting conditions and occlusions. A result from our pipeline is showcased in Fig.~\ref{fig:teaser}. 

The seminal work in the field is the \threed{} Morphable Model (\threedmm{}) fitting algorithm~\cite{blanz1999morphable}. 
The facial texture and shape that is reconstructed by the \threedmm{} algorithm 
always lies in a space that is spanned by a linear basis 
which is learned by Principal Component Analysis (\textsc{pca}). 
The linear basis, 
even though remarkable in representing the basic characteristics of the reconstructed face, 
fails in reconstructing high-frequency details in texture and geometry. 
Furthermore, the \textsc{pca} model fails in representing the complex structure of facial texture captured ``in-the-wild''. 
Therefore, \threedmm{} fitting usually fails on ``in-the-wild'' images. Recently, \threedmm{} fitting has been extended so that it uses a \textsc{pca} model on robust features, 
i.e., Histogram of Oriented Gradients (HoGs) \cite{dalal2005histograms}, 
for representing facial texture \cite{booth20173d}. 
The method has shown remarkable results in reconstructing 
the \threed{} facial geometry from ``in-the-wild'' images. 
Nevertheless, it cannot reconstruct facial texture that accurately.

With the advent of deep learning, 
many regression methods using an encoder-decoder structure 
have been proposed to infer \threed{} geometry, reflectance and illumination \cite{chen2019photo, gecer2019ganfit,sela2017unrestricted,shu2017neural,tewari2018self,tran2019learning,wang2019adversarial,zhou2019dense}. 
Some of the methods demonstrate that it is possible to reconstruct shape and texture, 
even in real-time on a \textsc{cpu} \cite{zhou2019dense}. 
Nevertheless, due to various factors, 
such as the use of basic reflectance models (e.g., the Lambertian reflectance model), 
the use of synthetic data or mesh-convolutions on colored meshes, the methods 
\cite{sela2017unrestricted,shu2017neural,tewari2018self,tran2019learning,wang2019adversarial, zhou2019dense}
fail to reconstruct highly-detailed texture and shape that is \renderready{}. Furthermore, in many of the above methods the reconstructed texture and shape lose many of the identity characteristics of the original image.

Arguably, the first generic method that demonstrated 
that it is possible to reconstruct high-quality texture and shape 
from single ``in-the-wild'' images 
is the recently proposed \textsc{ganfit} method \cite{gecer2019ganfit}. 
\textsc{ganfit} can be described as an extension of the original \threedmm{} fitting strategy 
but with the following differences: 
(a) instead of a \textsc{pca} texture model, it uses a Generative Adversarial Network (\textsc{gan}) \cite{karras2017progressive} trained on large-scale high-resolution \textsc{uv}-maps, 
and (b) in order to preserve the identity in the reconstructed texture and shape, 
it uses features from a state-of-the-art face recognition network \cite{deng2019arcface}.
However, the reconstructed texture and shape is not \renderready{} due to 
(a) the texture containing baked illumination, 
and (b) not being able to reconstruct high-frequency normals or specular reflectance. 

Early attempts to infer photorealistic \renderready{} information 
from single ``in-the-wild'' images have been made in the line of research of
\cite{chen2019photo, huynh2018mesoscopic,saito2017photorealistic, yamaguchi2018high}. 
Arguably, some of the results showcased in the above noted papers are of high-quality.
Nevertheless, the methods do not generalize since: 
(a) they directly manipulate and augment 
the low-quality and potentially occluded input facial texture,
instead of reconstructing it,
and as a result,
the quality of the final reconstruction always depends on the input image.
(b) the employed \threed{} model is not very representative, 
and (c) a very small number of subjects (e.g., $25$ \cite{yamaguchi2018high}) were available 
for training for the high-frequency details of the face. 
Thus, while closest to our work, 
these approaches focus on easily creating a  digital avatar 
rather than high-quality \renderready{} face reconstruction 
from ``in-the-wild'' images which is the goal of our work. 

In this paper, we propose the first, 
to the best of our knowledge,
methodology that produces high-quality \renderready{} face reconstructions from arbitrary images. 
In particular, 
our method builds upon recent reconstruction methods (e.g.,~\textsc{ganfit} \cite{gecer2019ganfit}) 
and contrary to \cite{chen2019photo, yamaguchi2018high} 
does not apply algorithms for high-frequency estimation to the original input, 
which could be of very low quality,
but to a \textsc{gan}-generated high-quality texture. 
Using a light stage,
we have collected a large scale dataset with samples of over 200
subjects' reflectance and geometry
and we train image translation networks that can perform estimation of 
(a) diffuse and specular albedo, and 
(b) diffuse and specular normals. 
We demonstrate that it is possible to produce \renderready{} faces from arbitrary faces (pose, occlusion, etc.) including portraits and face sketches,
which can be realistically relighted in any environment.
\begin{figure*}[ht!]
  \includegraphics[width=\textwidth]{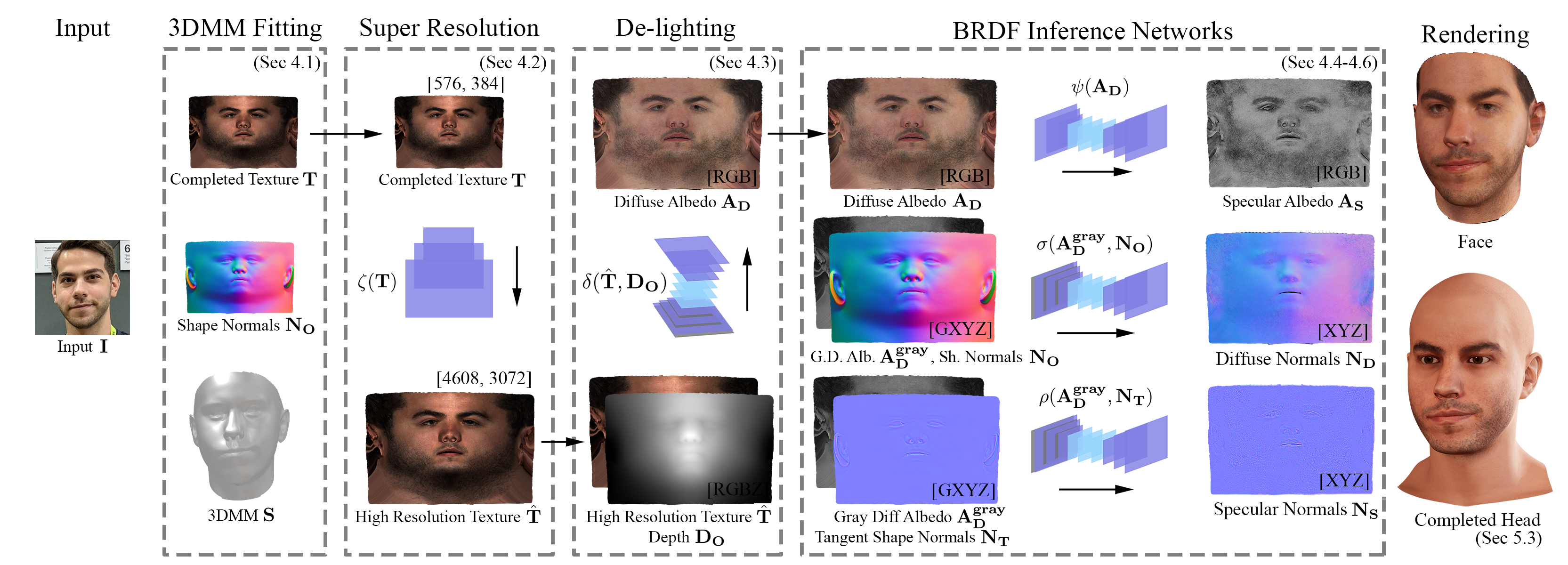}
  \caption{
    Overview of the proposed method.
    A \threedmm{} is fitted to an ``in-the-wild'' input image 
    and a completed \textsc{uv} texture is synthesized,
    while optimizing for the identity match between the rendering and the input.
    The texture is up-sampled $8$ times,
    to synthesize plausible high-frequency details.
    We then use an image translation network to de-light the texture
    and obtain the diffuse albedo with high-frequency details. 
    Then, separate networks infer the specular albedo,
    diffuse normals and specular normals (in tangent space) from the diffuse albedo and the \threedmm{} shape normals.
    The networks are trained on $512\times512$ patches and inferences are ran on $1536\times1536$ patches with a sliding window.
    Finally,
    we transfer the facial shape and consistently inferred reflectance
    to a head model.
    Both face and head can be rendered realistically in any environment.
  }
\label{fig:overview}
\end{figure*}

\section{Related Work}
\subsection{Facial Geometry and Reflectance Capture}

Debevec et al.~\cite{Debevec:2000} first proposed employing a specialized light stage setup to acquire a reflectance field of a human face for photo-realistic image-based relighting applications. They also employed the acquired data to estimate a few view-dependent reflectance maps for rendering. Weyrich et al.~\cite{Weyrich:2006} employed an LED sphere and 16 cameras to densely record facial reflectance and computed view-independent estimates of facial reflectance from the acquired data including per-pixel diffuse and specular albedos, and per-region specular roughness parameters. These initial works employed dense capture of facial reflectance which is somewhat cumbersome and impractical.

Ma et al.~\cite{ma2007rapid} introduced polarized spherical gradient illumination (using an LED sphere) for efficient acquisition of separated diffuse and specular albedos and photometric normals of a face using just eight photographs, and demonstrated high quality facial geometry, including skin mesostructure as well as realistic rendering with the acquired data. It was however restricted to a frontal viewpoint of acquisition due to their employment of view-dependent polarization pattern on the LED sphere. Subsequently, Ghosh et al.~\cite{ghosh2011multiview} extended polarized spherical gradient illumination for multi-view facial acquisition by employing two orthogonal spherical polarization patterns. Their method allows capture of separated diffuse and specular reflectance and photometric normals from any viewpoint around the equator of the LED sphere and can be considered the state-of-the art in terms of high quality facial capture.

Recently, Kampouris et al.~\cite{kampouris2018diffuse} demonstrated how to employ unpolarized binary spherical gradient illumination for estimating separated diffuse and specular albedo and photometric normals using color-space analysis. The method has the advantage of not requiring polarization and hence requires half the number of photographs compared to polarized spherical gradients and enables completely view-independent reflectance separation, making it faster and more robust for high quality facial capture~\cite{lattas2019multi}.

Passive multiview facial capture has also made significant progress in recent years, from high quality facial geometry capture~\cite{Beeler:2010} to even detailed facial appearance estimation~\cite{Gotardo:2018}. However, the quality of the acquired data with such passive capture methods is somewhat lower compared to active illumination techniques.

In this work, we employ two state-of-the-art active illumination based multiview facial capture methods~\cite{ghosh2011multiview,lattas2019multi} for acquiring high quality facial reflectance data in order to build our training data.

\subsection{Image-to-Image Translation}
Image-to-image translation refers to the task of translating an input image 
to a designated target domain (e.g., turning sketches into images, or day into night scenes). 
With the introduction of \textsc{gan}s \cite{goodfellow2014generative}, 
image-to-image translation improved dramatically \cite{isola2017image, zhu2017unpaired}. Recently, with the increasing capabilities in the hardware, 
image-to-image translation has also been successfully attempted 
in high-resolution data \cite{wang2018high}. 
In this work we utilize variations of pix2pixHD~\cite{wang2018high} 
to carry out tasks such as de-lighting and the extraction of reflectance maps in very high-resolution.

\subsection{Facial Geometry Estimation}
Over the years, numerous methods have been introduced in the literature 
that tackle the problem of \threed{} facial reconstruction from a single input image. 
Early methods required a statistical \threedmm{}
both for shape and appearance,
usually encoded in a low dimensional space constructed by \textsc{pca} \cite{blanz1999morphable,booth20173d}. 
Lately, many approaches have tried to leverage the power of Convolutional Neural Networks (\textsc{cnn}s) 
to either regress the latent parameters of a \textsc{pca} model \cite{tuan2017regressing,cole2017synthesizing} 
or utilize a \threedmm{} to synthesize images 
and formulate an image-to-image translation problem using \textsc{cnn}s \cite{guo2018cnn, richardson20163d}.

\subsection{Photorealistic 3D faces with Deep Learning}
Many approaches have been successful in acquiring the reflectance of materials from a single image,
using deep networks with an encoder-decoder architecture
\cite{deschaintre2018single,li2017modeling,li2018materials}.
However, they only explore \twod{} surfaces and in a constrained environment,
usually assuming a single point-light source.

Early applications on human faces \cite{sengupta2018sfsnet, shu2017neural}
used image translation networks to infer facial reflection from an ``in-the-wild'' image,
producing low-resolution results.
Recent approaches attempt to incorporate additional facial normal and displacement mappings
resulting in representations with high frequency details \cite{chen2019photo}.
Although this method demonstrates impressive results in geometry inference,
it tends to fail in conditions with harsh illumination and extreme head poses, 
and does not produce re-lightable results.
Saito et al.~\cite{saito2017photorealistic} proposed a deep learning approach 
for data-driven inference of high resolution facial texture map of an entire face 
for realistic rendering,
using an input of a single low-resolution face image with partial facial coverage.
This has been extended to inference of facial mesostructure, given a diffuse albedo
\cite{huynh2018mesoscopic},
and even complete facial reflectance and displacement maps besides albedo texture, given partial facial image as input~\cite{yamaguchi2018high}. 
While closest to our work, these approaches achieve the creation of digital avatars,
rather than high quality facial appearance estimation from ``in-the-wild'' images.
In this work, we try to overcome these limitations by employing an iterative optimization framework as proposed in~\cite{gecer2019ganfit}. 
This optimization strategy leverages a deep face recognition network and \textsc{gan}s into a conventional fitting method in order to estimate the high-quality geometry and texture with fine identity characteristics,
which can then be used to produce high-quality reflectance maps.

\section{Training Data}
\label{sec:dataset}

\subsection{Ground Truth Acquisition}

\begin{figure}[h]
\vspace{-0.3cm}
\begin{subfigure}{.24\linewidth}
  \centering
  \includegraphics[width=\linewidth]{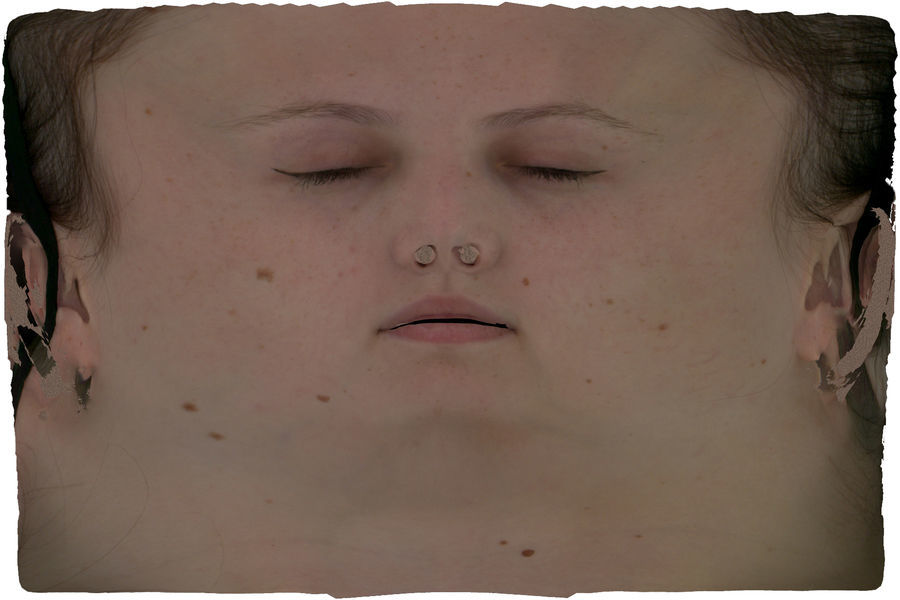}
  \includegraphics[width=\linewidth]{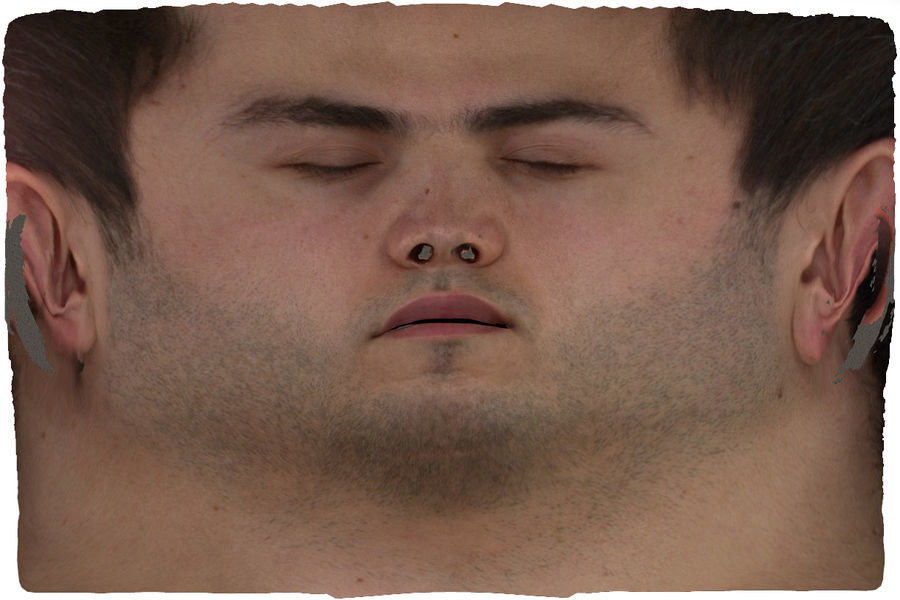}
  \caption{Diff. Alb.}
  \label{fig:4_data_diffAlbedo}
\end{subfigure}%
\vspace{1pt}
\begin{subfigure}{.24\linewidth}
  \centering
  \includegraphics[width=\linewidth]{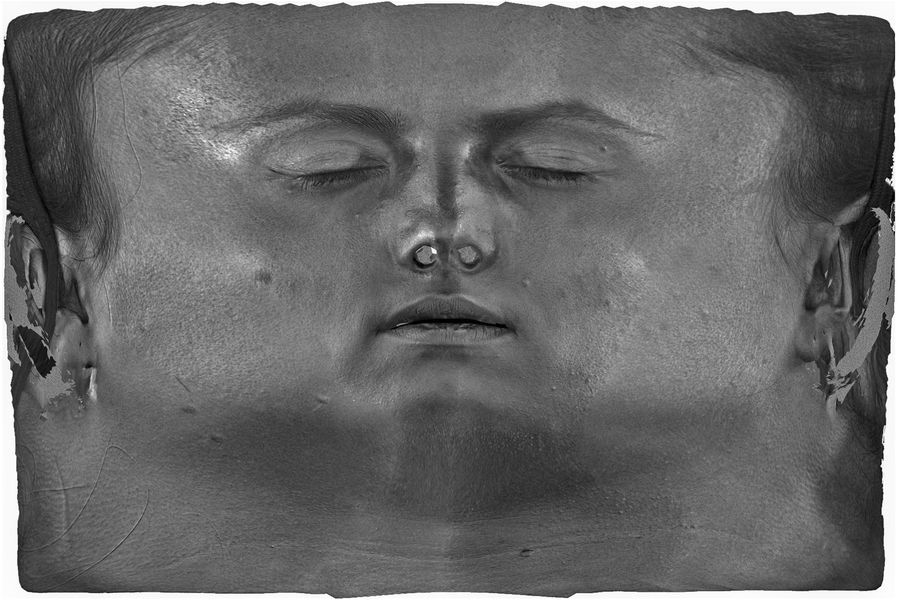}
  \includegraphics[width=\linewidth]{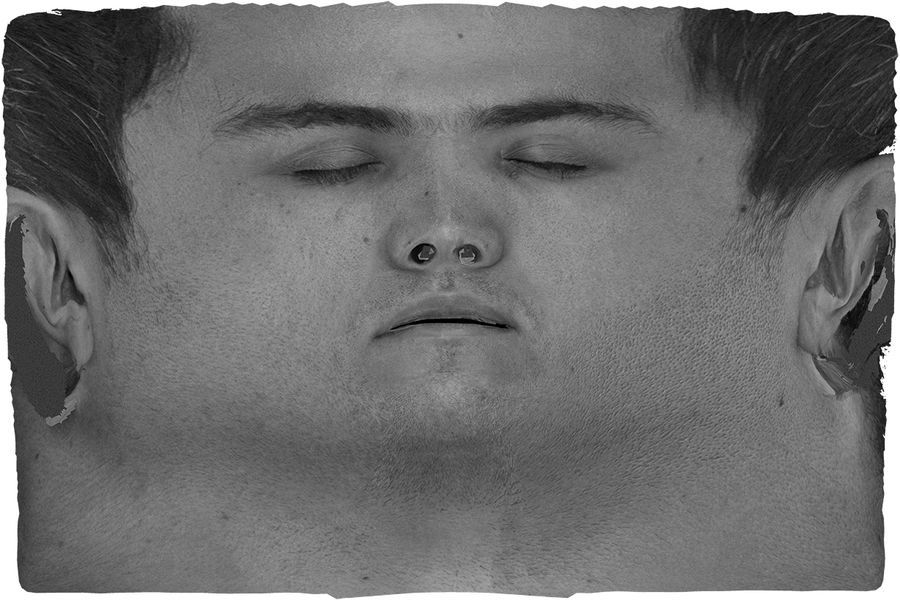}
  \caption{Spec. Alb.}
  \label{fig:4_data_diffNormals}
\end{subfigure}
\begin{subfigure}{.24\linewidth}
  \centering
  \includegraphics[width=\linewidth]{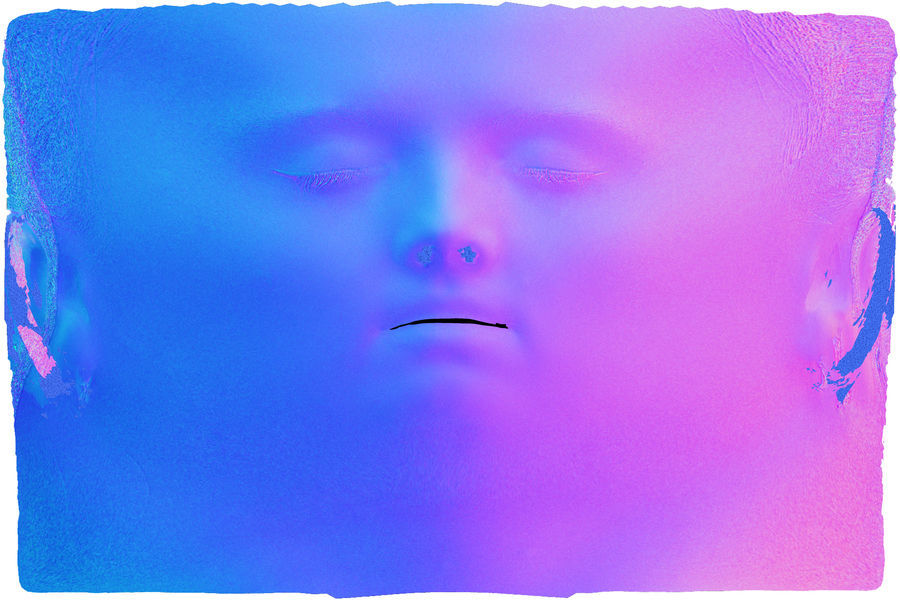}
  \includegraphics[width=\linewidth]{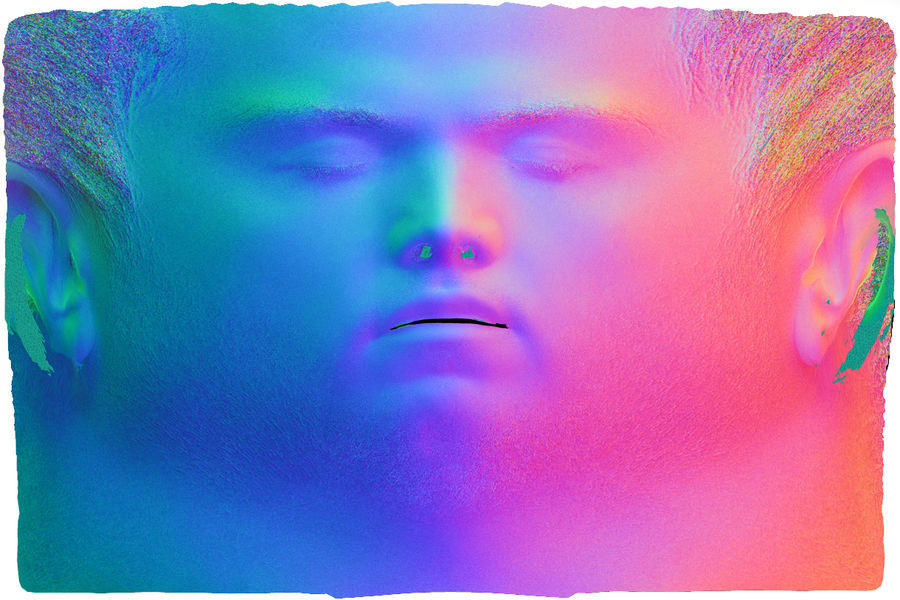}
  \caption{Diff. Nor.}
  \label{fig:4_data_diffNormals}
\end{subfigure}
\begin{subfigure}{.24\linewidth}
  \centering
  \includegraphics[width=\linewidth]{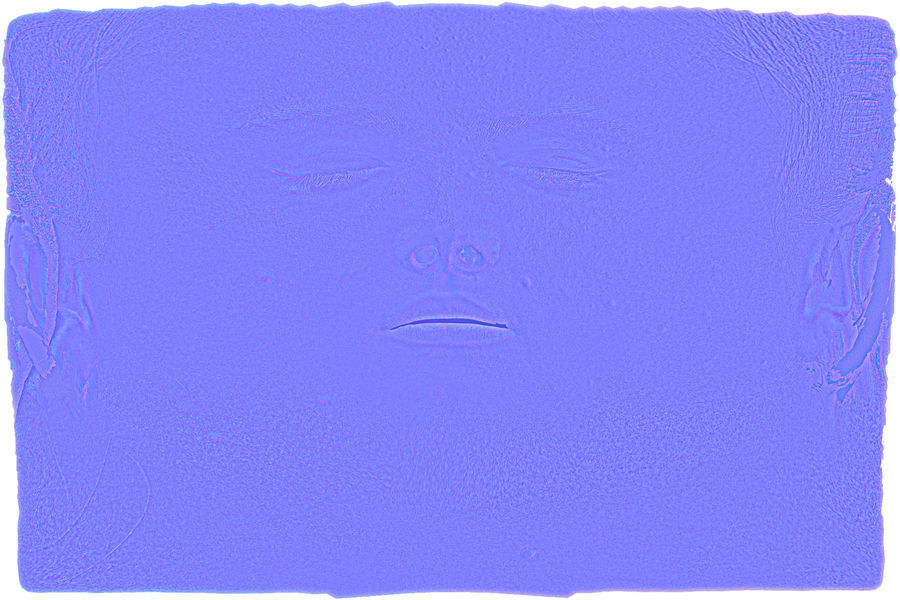}
  \includegraphics[width=\linewidth]{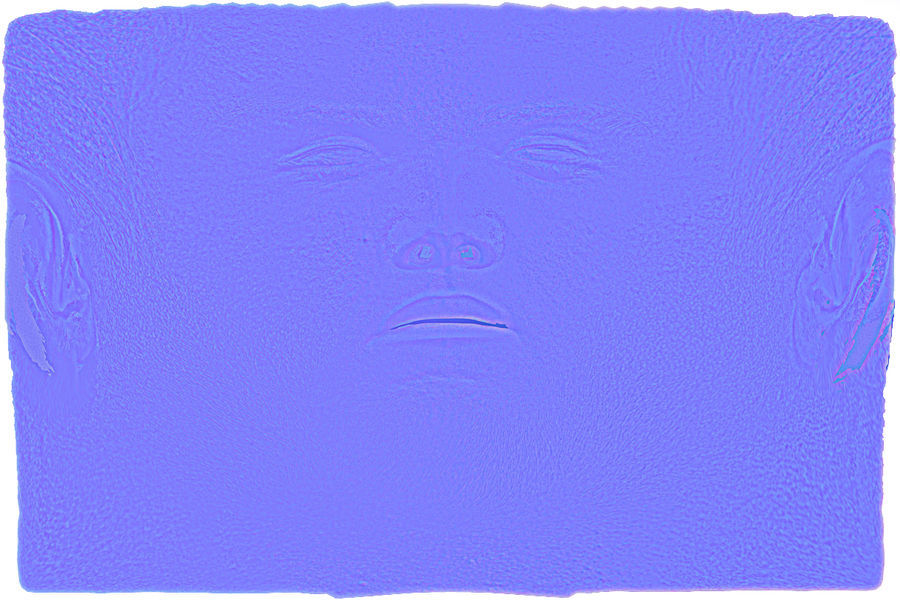}
  \caption{Spec. Nor.}
  \label{fig:4_data_diffNormals}
\end{subfigure}
\caption{Two subjects' reflectance acquired with \cite{ghosh2011multiview} (top) and \cite{kampouris2018diffuse, lattas2019multi} (bottom).
Specular normals in tangent space.
}
\label{fig:4_data_capturedData}
\end{figure}

We employ the state-of-the-art method of \cite{ghosh2011multiview} for capturing high resolution pore-level reflectance maps of faces using a polarized \textsc{led} sphere with 168 lights (partitioned into two polarization banks) and 9 \textsc{dslr} cameras. Half the \textsc{led}s on the sphere are vertically polarized (for parallel polarization), and the other half are horizontally polarized (for cross-polarization) in an interleaved pattern. 

Using the \textsc{led} sphere, we can also employ the color-space analysis from unpolarised \textsc{led}s \cite{kampouris2018diffuse} for diffuse-specular separation and the multi-view facial capture method of \cite{lattas2019multi}
to acquire unwrapped textures of similar quality (Fig. \ref{fig:4_data_capturedData}).
This method requires less than half of data captured (hence reduced capture time) and a simpler setup (no polarizers),
enabling the acquisition of larger datasets.

\subsection{Data Collection}
In this work, we capture faces of over 200 individuals 
of different ages and characteristics under 7 different expressions. 
The geometry reconstructions are registered to a standard topology,
like in \cite{booth20163d}, with unwrapped textures as shown in Fig. \ref{fig:4_data_capturedData}.
We name the dataset \datasetname.
It is currently the largest dataset of this type and we intend to make it publicly available to the scientific community 
\footnote{
For the dataset and other materials we refer the reader to the project's page
\url{https://github.com/lattas/avatarme}.
}.

\section{Method}\label{sec:method}

\begin{figure}[ht]
\begin{subfigure}{.19\linewidth}
  \centering
  \includegraphics[width=\linewidth]{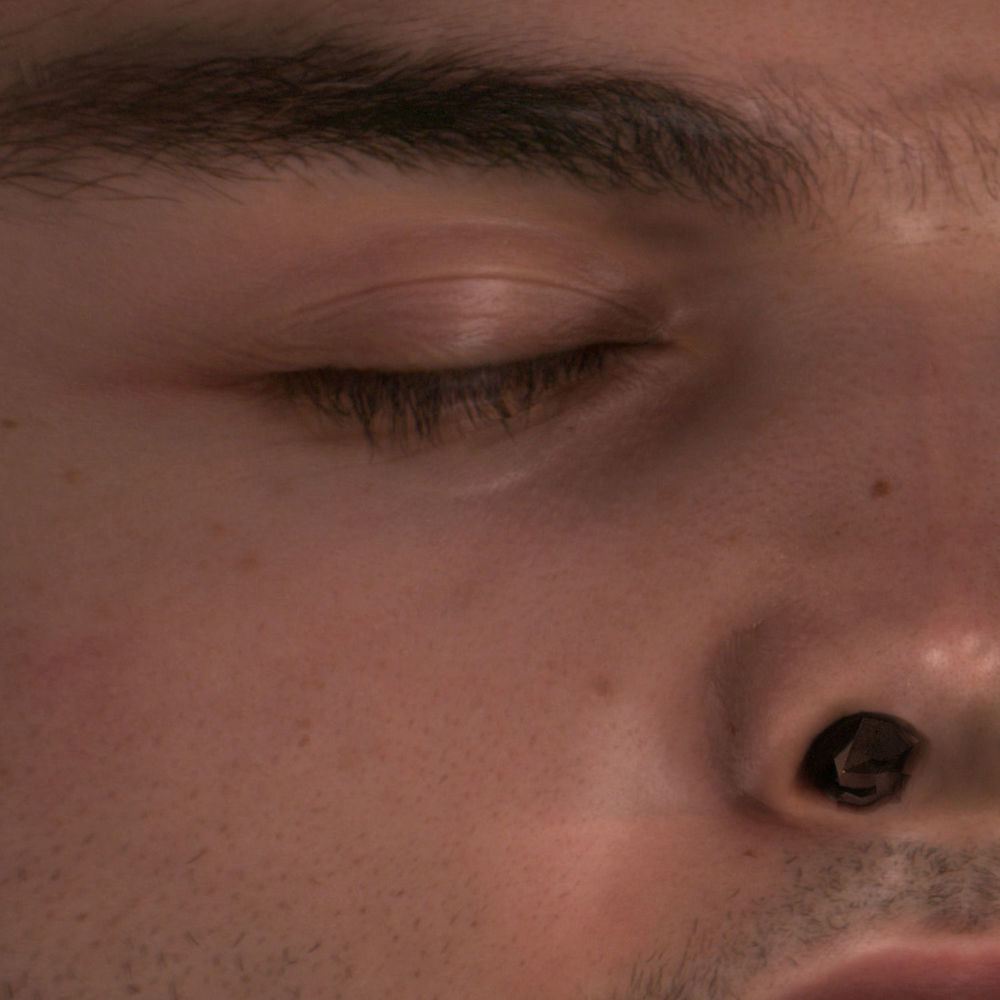}
  \caption{Input}
  \label{fig:5_method_render}
\end{subfigure}%
\vspace{1pt}
\begin{subfigure}{.19\linewidth}
  \centering
  \includegraphics[width=\linewidth]{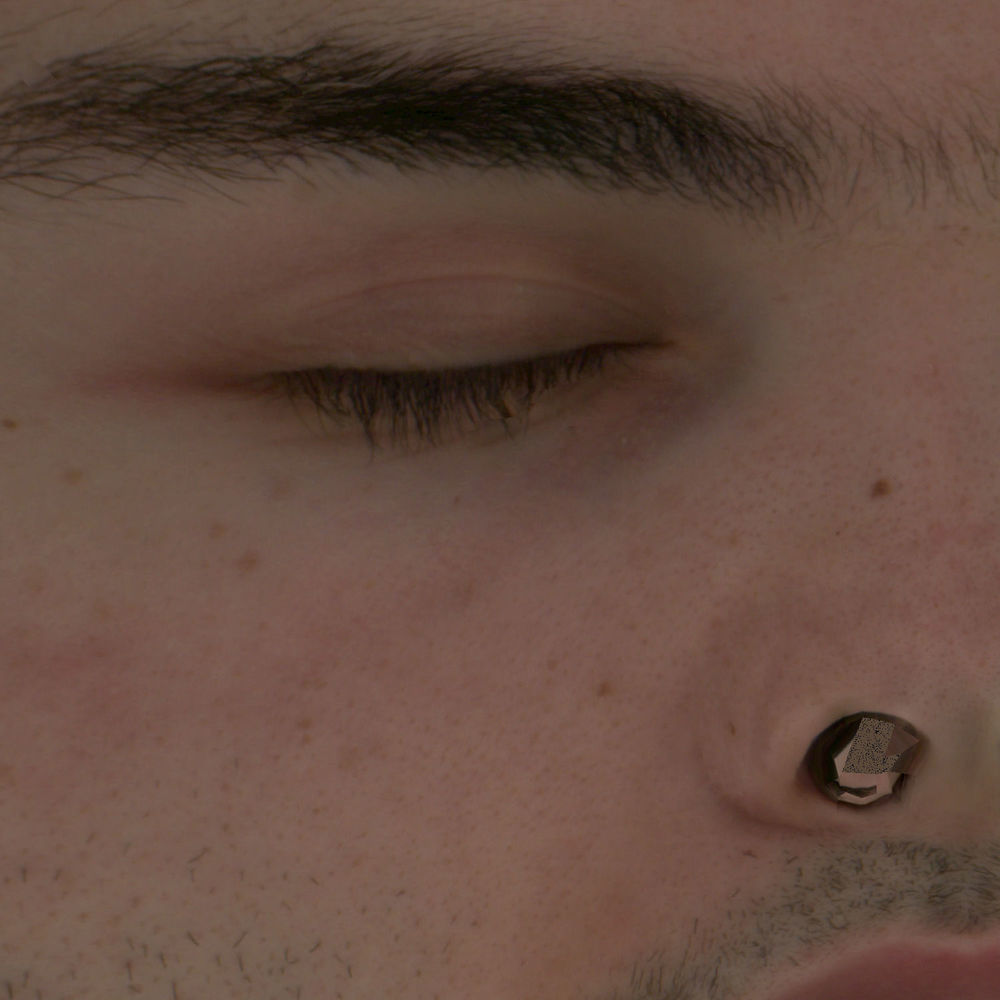}
  \hspace{1pt}
  \includegraphics[width=\linewidth]{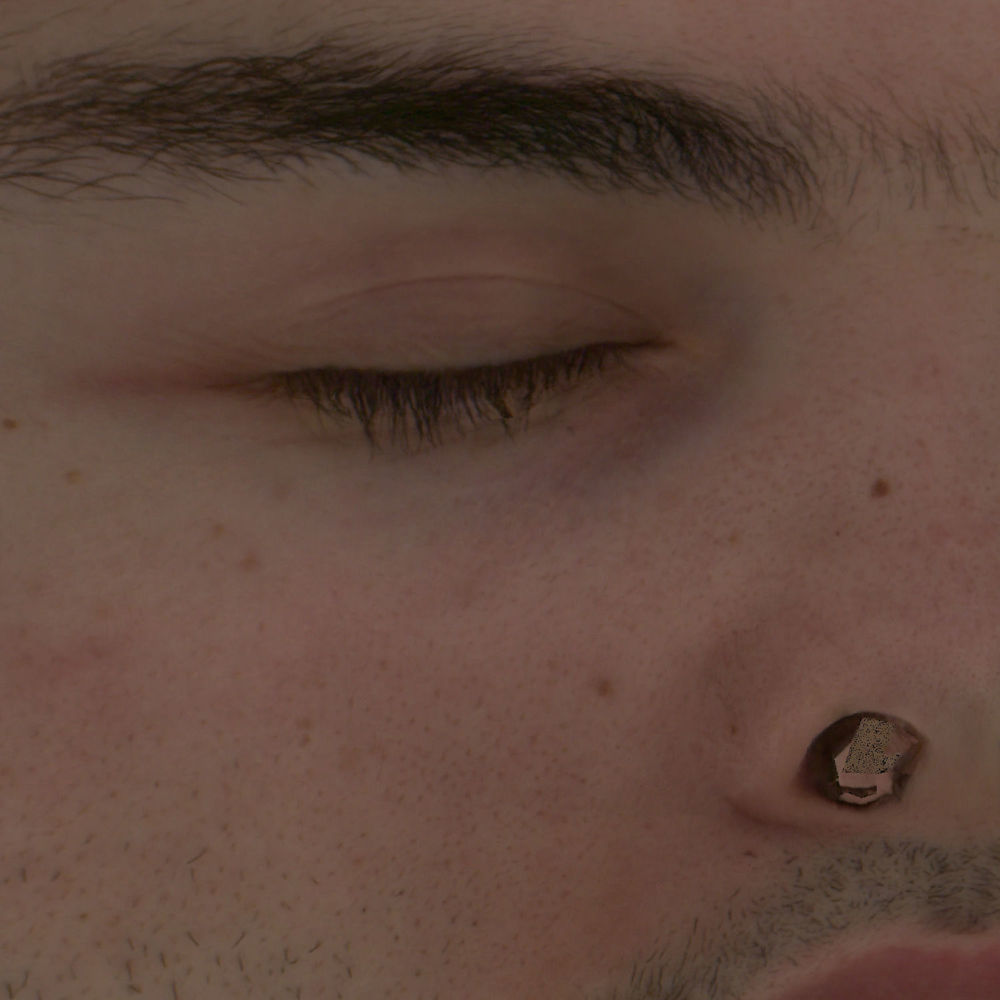}
  \caption{Diff Alb}
  \label{fig:4_method_diffAlbedo}
\end{subfigure}
\begin{subfigure}{.19\linewidth}
  \centering
  \includegraphics[width=\linewidth]{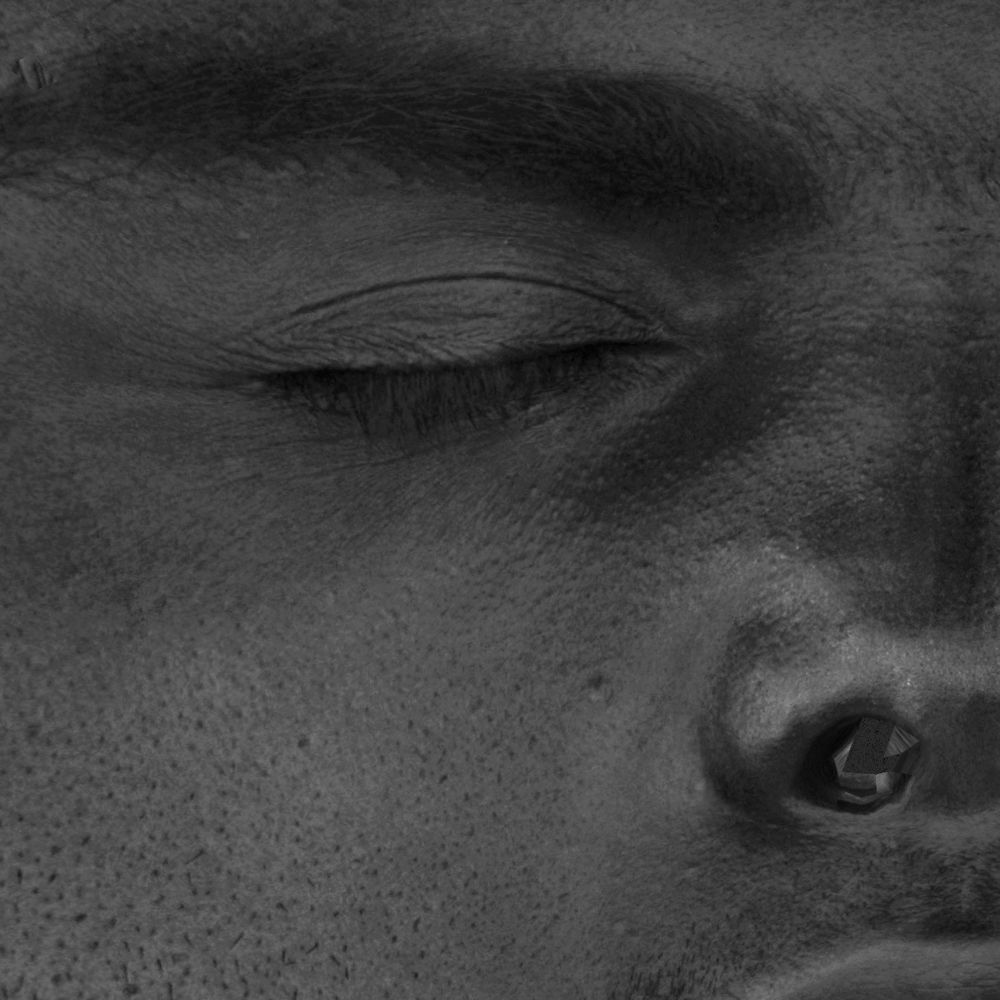}
  \vspace{1pt}
  \includegraphics[width=\linewidth]{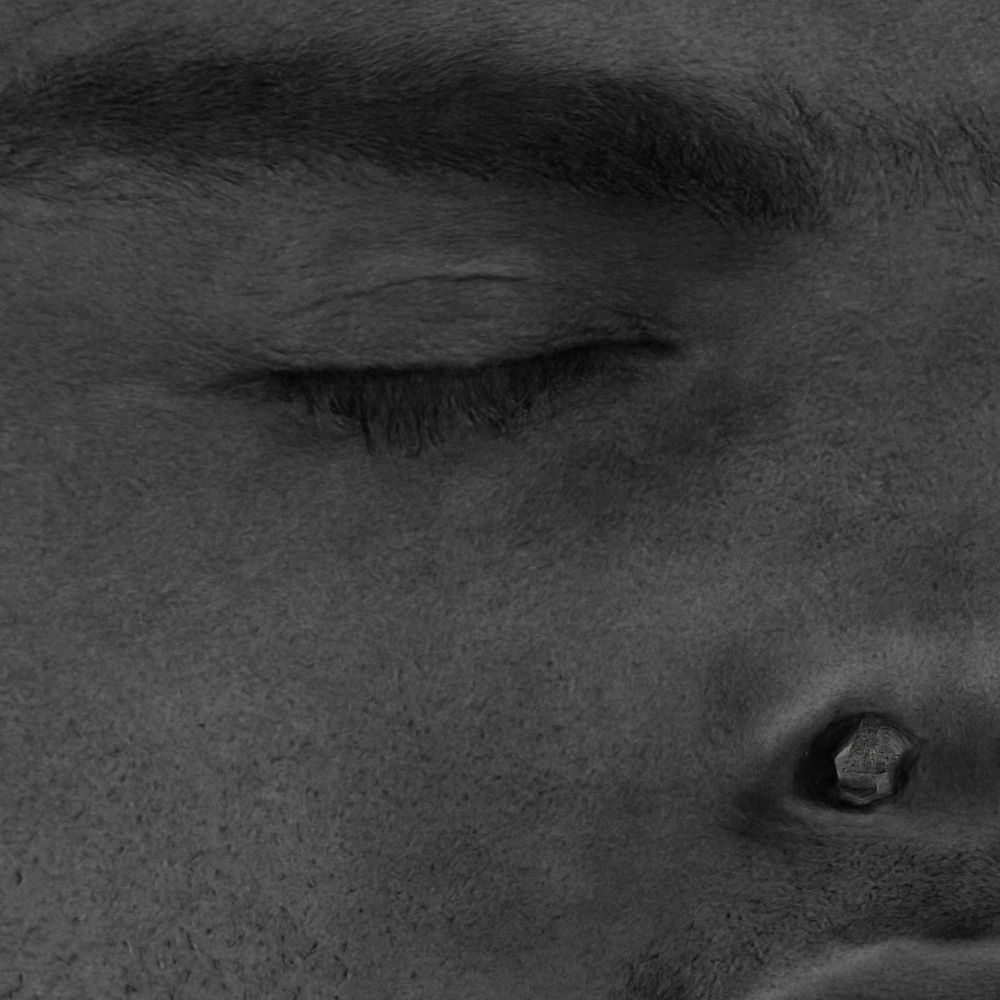}
  \caption{Spec Alb}
  \label{fig:5_method_specAlbedo}
\end{subfigure}
\begin{subfigure}{.19\linewidth}
  \centering
  \includegraphics[width=\linewidth]{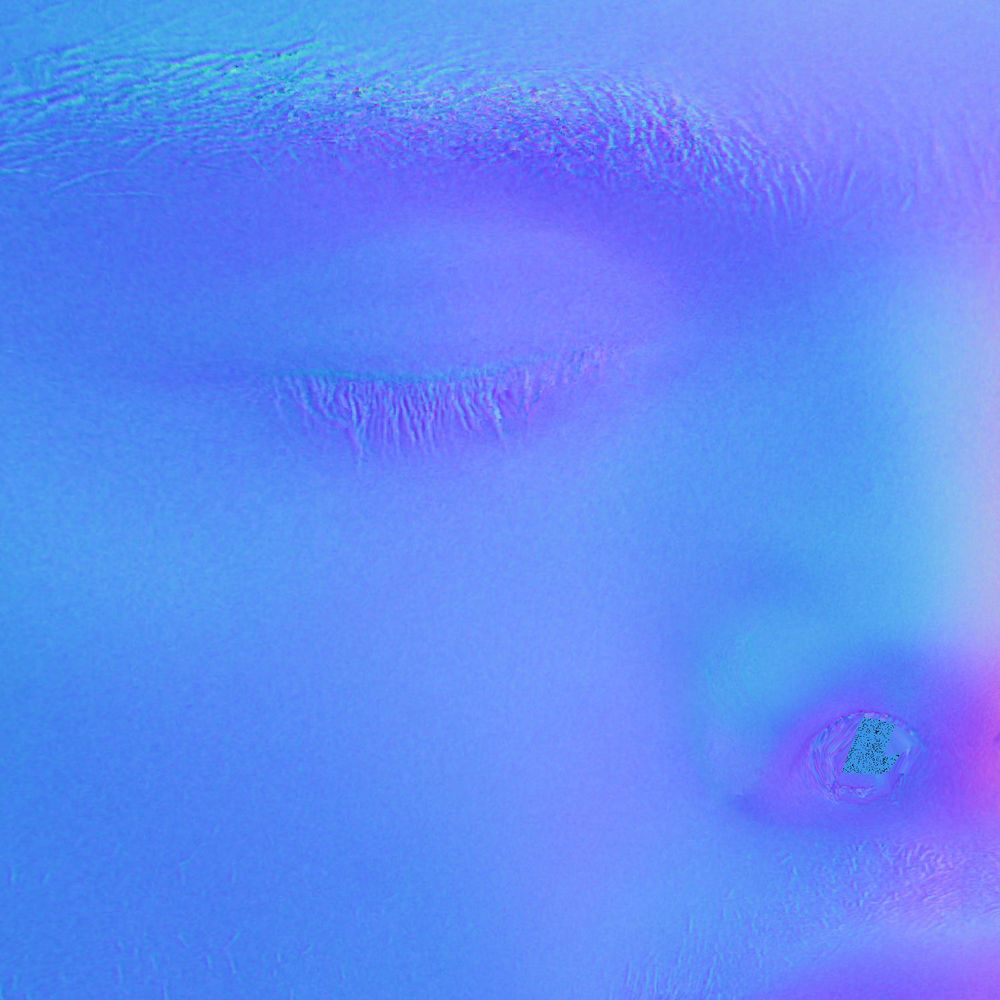}
  \vspace{1pt}
  \includegraphics[width=\linewidth]{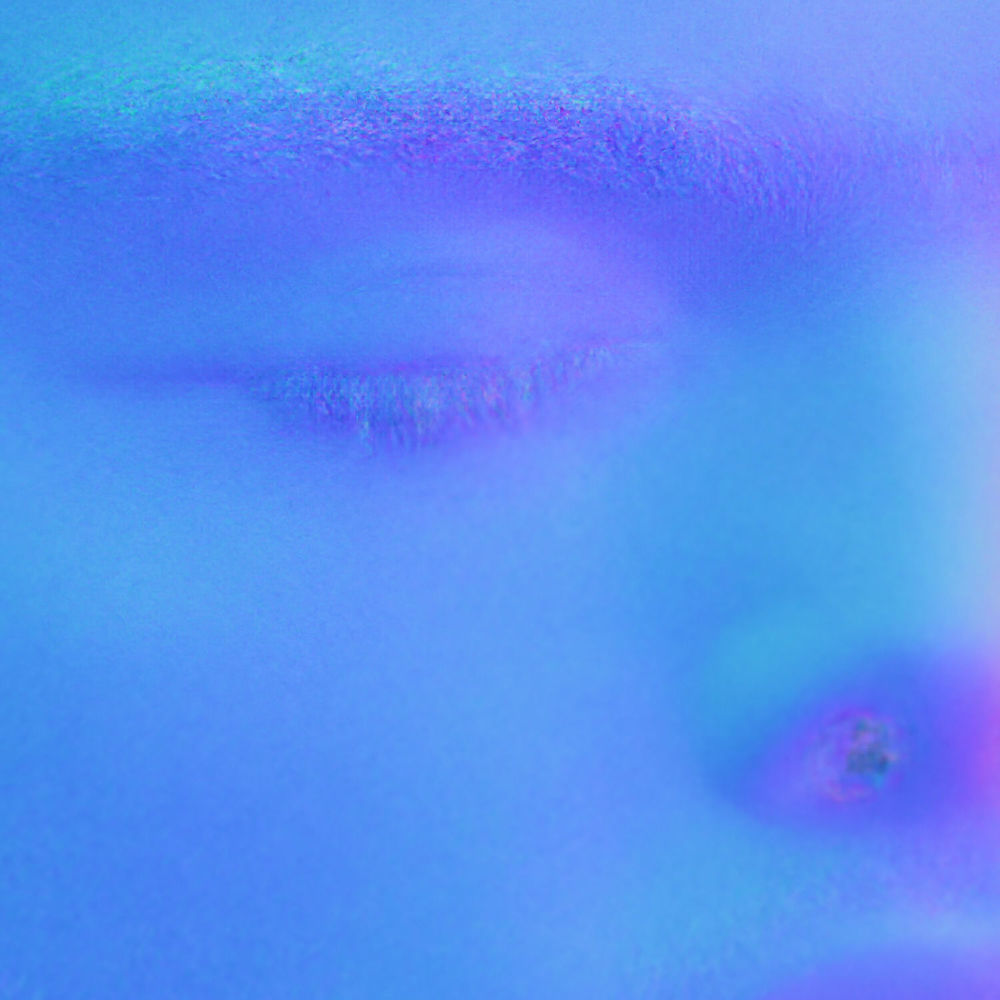}
  \caption{Diff Nor}
  \label{fig:5_method_specAlbedo}
\end{subfigure}
\begin{subfigure}{.19\linewidth}
  \centering
  \includegraphics[width=\linewidth]{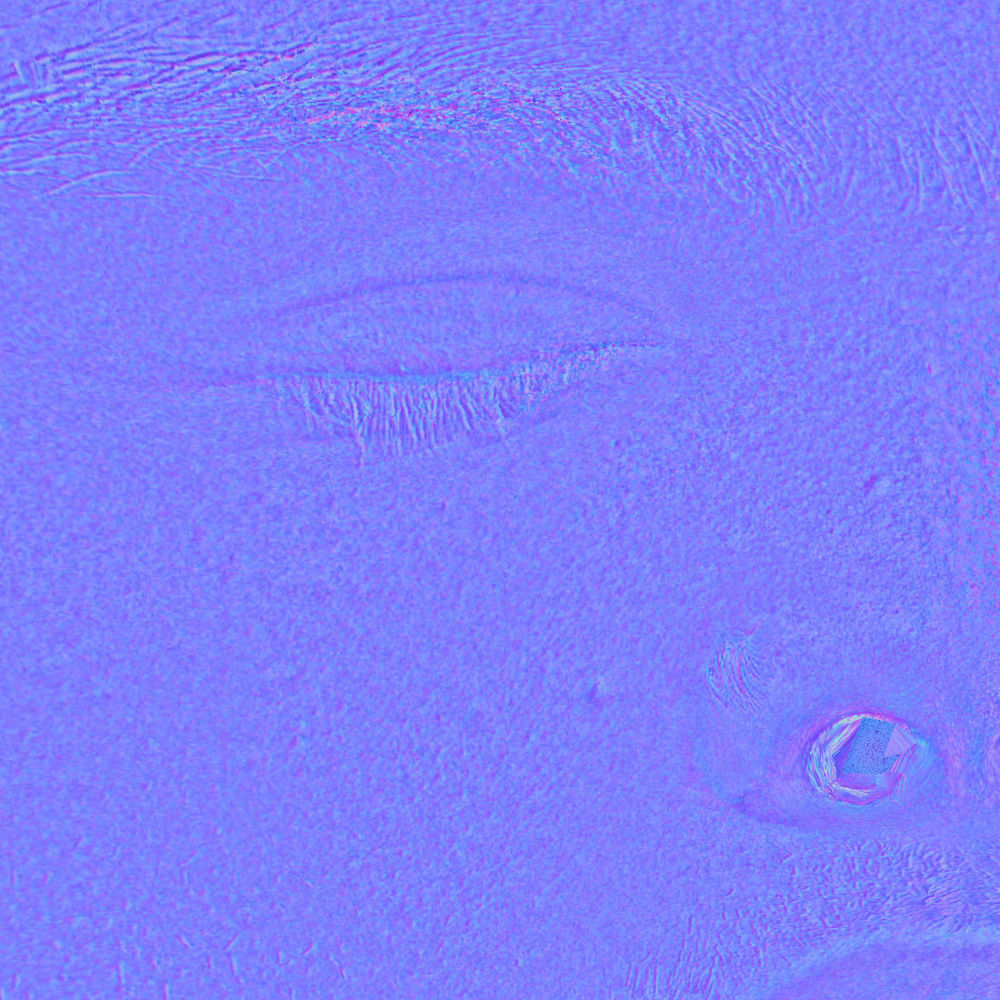}
  \vspace{1pt}
  \includegraphics[width=\linewidth]{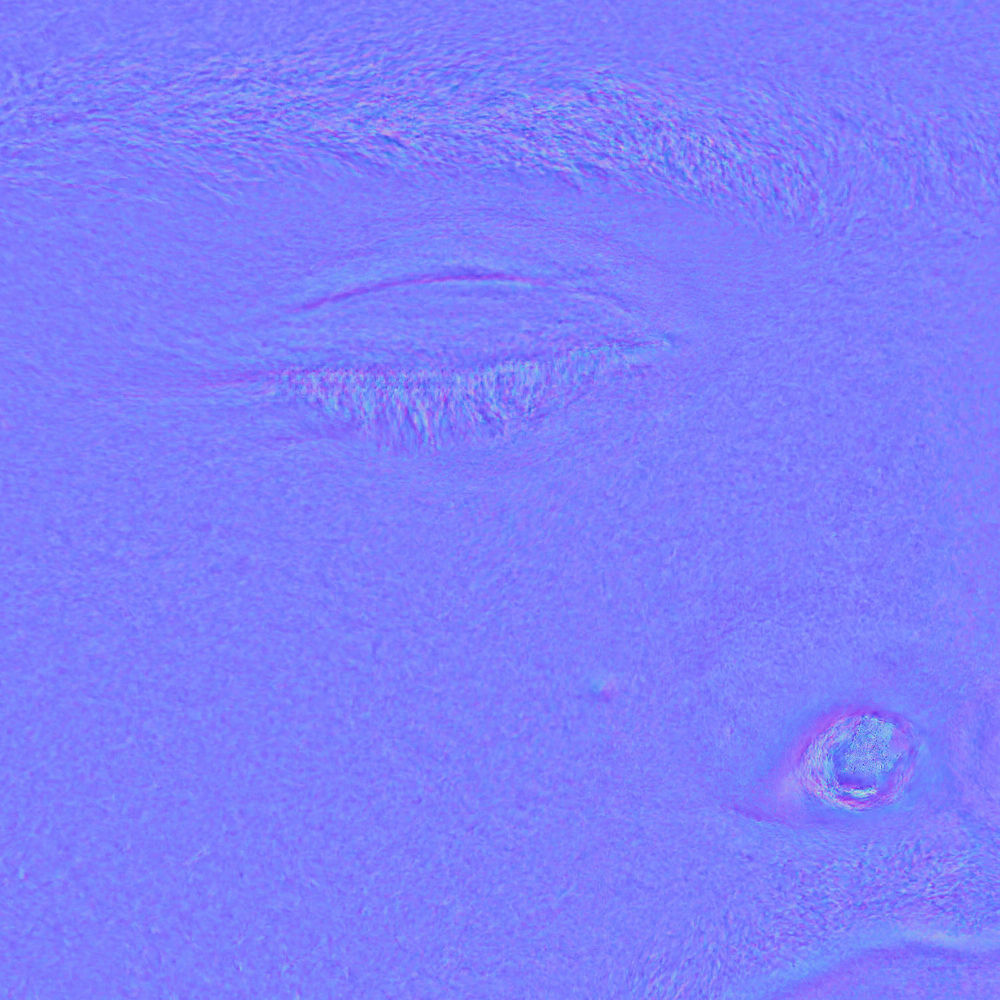}
  \caption{Spec Nor}
  \label{fig:5_method_tangentNormals}
\end{subfigure}
\caption{
    Rendered patch (\cite{gecer2019ganfit}-like) of a subject acquired with \cite{ghosh2011multiview},
    ground truth maps (top-row) and predictions with our network given rendering as input (bottom-row).
}
\label{fig:5_method_results}
\end{figure}

To achieve photorealistic rendering of the human skin,
we separately model the diffuse and specular albedo and normals
of the desired geometry.
Therefore, given a single unconstrained face image as input, 
we infer the facial geometry as well as the
\textit{diffuse albedo} ($\mathbf{A_D}$), 
\textit{diffuse normals} ($\mathbf{N_D}$)~\footnote{
    The diffuse normals $\mathbf{N_D}$ are not usually used in commercial rendering systems. By inferring $\mathbf{N_D}$ we can model the reflection as in the state-of-the-art specular-diffuse separation techniques \cite{ghosh2011multiview, lattas2019multi}.},
\textit{specular albedo} ($\mathbf{A_S})$, 
and \textit{specular normals} ($\mathbf{N_S}$).

As seen in Fig.~\ref{fig:overview}, we first reconstruct a \threed{} face (base geometry with texture) 
from a single image at a low resolution using an existing \threedmm{} algorithm \cite{booth20163d}. 
Then, the reconstructed texture map, which contains baked illumination,
is enhanced by a super resolution network,
followed by a de-lighting network to obtain a high resolution diffuse albedo $\mathbf{A_D}$.
Finally, we infer the other three components 
($\mathbf{A_S}, \mathbf{N_D}, \mathbf{N_S}$)
from the diffuse albedo $\mathbf{A_D}$ in conjunction with the base geometry. The following sections explain these steps in detail.

\subsection{Initial Geometry and Texture Estimation}
Our method requires a low-resolution \threed{} reconstruction 
of a given face image $\mathbf{I}$. 
Therefore, we begin with the estimation of the facial shape with $n$ vertices 
$\mathbf{S} \in \mathbb{R}^{n\times3}$
and texture
$\mathbf{T} \in \mathbb{R}^{576\times384\times3}$ 
by borrowing any state-of-the-art \threed{} face reconstruction approach (we use  \textsc{ganfit}~\cite{gecer2019ganfit}). 
Apart from the usage of deep identity features, 
\textsc{ganfit} synthesizes realistic texture \textsc{uv} maps 
using a \textsc{gan} as a statistical representation of the facial texture.
We reconstruct the initial base shape and texture of the input image $\mathbf{I}$ as follows
and refer the reader to \cite{gecer2019ganfit} for further details:
\begin{align}
    \mathbf{T}, \mathbf{S} = \mathcal{G}(\mathbf{I})
\end{align}

\noindent where 
$\mathcal{G}:  \mathbb{R}^{k \times m \times 3} \mapsto \mathbb{R}^{576\times384\times3}, \mathbb{R}^{n\times3}$ 
denotes the \textsc{ganfit} reconstruction method for an 
$\mathbb{R}^{k \times m \times 3}$ arbitrary sized image,
and $n$ number of vertices on a fixed topology.

Having acquired the prerequisites,
we procedurally improve on them:
from the reconstructed geometry $\mathbf{S}$, we acquire the shape normals $\mathbf{N}$ 
and enhance the facial texture $\mathbf{T}$ resolution,
before using them to estimate the components for physically based rendering,
such as the diffuse and specular diffuse and normals.
\subsection{Super-resolution}
Although the texture 
$\mathbf{T} \in \mathbb{R}^{576\times384\times3}$ 
from \textsc{ganfit} \cite{gecer2019ganfit} has reasonably good quality, 
it is \textit{below par} compared to artist-made \renderready{} \threed{} faces.
To remedy that, we employ a state-of-the-art super-resolution network,
\textsc{rcan}~\cite{zhang2018image}, 
to increase the resolution of the \textsc{uv} maps from 
$\mathbf{T} \in \mathbb{R}^{576\times384\times3}$ to 
$\hat{\mathbf{T}} \in \mathbb{R}^{4608\times3072\times3}$,
which is then retopologized and up-sampled to $\mathbb{R}^{6144\times4096}$.
Specifically, we train a super-resolution network
$(\zeta:  \mathbb{R}^{48\times48\times3} \mapsto \mathbb{R}^{384\times384\times3})$ 
with the texture patches of the acquired low-resolution texture $\mathbf{T}$.
At the test time, the whole texture from \textsc{ganfit} $\mathbf{T}$ 
is upscaled by the following:
\begin{align}
    \hat{\mathbf{T}} = \zeta(\mathbf{T})
\end{align}

\subsection{Diffuse Albedo Extraction by De-lighting}
A significant issue of the texture $\mathbf{T}$ produced by \threedmm{}s
is that they are trained on data with baked illumination (i.e. reflection, shadows), which they reproduce.
\textsc{ganfit}-produced textures contain sharp highlights and shadows,
made by strong point-light sources, as well as baked environment illumination,
which prohibits photorealistic rendering.
In order to alleviate this problem, 
we first model the illumination conditions of the dataset used in \cite{gecer2019ganfit}
and then synthesize UV maps with the same illumination 
in order to train an image-to-image translation network from texture with baked-illumination 
to unlit diffuse albedo $\mathbf{A_D}$. 
Further details are explained in the following sections.

\subsubsection{Simulating Baked Illumination}
\label{sec:simulation}
Firstly, we acquire random texture and mesh outputs from \textsc{ganfit}.
Using a cornea model \cite{nishino2004eyes}, 
we estimate the average direction of the apparent 3 point light sources used,
with respect to the subject, 
and an environment map for the textures $\mathbf{T}$. 
The environment map produces a good estimation of the environment illumination of \textsc{ganfit}'s data
while the 3 light sources help to simulate the highlights and shadows.
Thus, we render our acquired 200 subjects (Section \ref{sec:dataset}), 
as if they were samples from the dataset used in the training of \cite{gecer2019ganfit}, 
while also having accurate ground truth of their albedo and normals. 
We compute a physically-based rendering for each subject from all view-points,
using the predicted environment map and the predicted light sources 
with a random variation of their position,
creating an illuminated texture map. 
We denote this whole simulation process by  
$ \xi: \mathbf{A_D} \in  \mathbb{R}^{6144 \times 4096 \times 3} \mapsto \mathbf{A_D^T} \in \mathbb{R}^{6144\times4096\times3}$ 
which translates diffuse albedo to the distribution of the textures with baked illumination, as shown in the following:
\begin{align}
\mathbf{A_D^T} = \xi (\mathbf{A_D}) \sim \mathbb{E}_{\mathbf{t} \in \{\mathbf{T_1}, \mathbf{T_2}, \dots, \mathbf{T_n}\}} \mathbf{t}
\end{align}

\subsubsection{Training the De-lighting Network}
Given the simulated illumination as explained in Sec. \ref{sec:simulation},
we now have access to a version of \datasetname{} with the \cite{gecer2019ganfit}-like illumination $\mathbf{A_D^T}$
and with the corresponding diffuse albedo $\mathbf{A_D}$. 
We formulate de-lighting as a domain adaptation problem and train an image-to-image translation network. 
To do this, we follow two strategies different from the standard image translation approaches.

Firstly, we find that the occlusion of illumination on the skin surface
is geometry-dependent 
and thus the resulting albedo improves in quality when feeding the network
with both the texture and geometry of the \threedmm{}. 
To do so, we simply normalize the texture $\mathbf{A_D^T}$ channels to $[-1, 1]$ 
and concatenate them with the depth of the mesh in object space $\mathbf{D_O}$, also in $[-1, 1]$.
The depth ($\mathbf{D_O}$) is defined as the $Z$ dimension of the vertices
of the acquired and aligned geometries, in a \textsc{uv} map.
We feed the network with a \fourd{} tensor of
$[\mathbf{A^T_{D_R}},\mathbf{A^T_{D_G}},\mathbf{A^T_{D_B}},\mathbf{D_O}]$
and predict the resulting 3-channel albedo
$[\mathbf{A_{D_R}},\mathbf{A_{D_G}},\mathbf{A_{D_B}}]$.
Alternatively, we can also use as an input the texture $ \mathbf{A_D^T}$
concatenated with the normals in object space ($\mathbf{N_O}$).
We found that feeding the network only with the texture map
causes artifacts in the inference.
Secondly, we split the original high resolution data into overlapping patches of
$512 \times 512$ pixels in order to augment the number of data samples
and avoid overfitting. 

In order to remove existing illumination from $\hat{\mathbf{T}}$, 
we train an image-to-image translation network with patches $\delta:
\mathbf{A_D^T}, \mathbf{D_O} \mapsto \mathbf{A_D} \in
\mathbb{R}^{512\times512\times3}$ 
and then extract the diffuse albedo $\mathbf{A_D}$ by the following:
\begin{align}
\mathbf{A_D} = \delta(\hat{\mathbf{T}}, \mathbf{D_O})
\end{align}

\subsection{Specular Albedo Extraction}
\paragraph{Background:}
Predicting the entire specular \textsc{brdf}
and the per-pixel specular roughness from the illuminated texture $\mathbf{\hat{T}}$ 
or the inferred diffuse albedo $\mathbf{A_D}$, 
poses an unnecessary challenge.
As shown in \cite{ghosh2011multiview, kampouris2018diffuse}
a subject can be realistically rendered 
using only the intensity of the specular reflection $\mathbf{A_S}$,
which is consistent on a face due to the skin’s refractive index. 
The spatial variation is correlated to facial skin structures such as skin pores, wrinkles or hair, 
which act as reflection occlusions reducing the specular intensity.

\paragraph{Methodology:} In principle, 
the specular albedo can also be computed from the texture with the baked illumination, 
since the texture includes baked specular reflection. 
However, we empirically found that the specular component is strongly biased 
due to the environment illumination and occlusion. 
Having computed a high quality diffuse albedo $\mathbf{A_D}$ from the previous step, 
we infer the specular albedo $\mathbf{A_S}$
by a similar patch-based image-to-image translation network 
from the diffuse albedo
($\psi: \mathbf{A_D} \mapsto \mathbf{A_S} \in
\mathbb{R}^{512\times512\times3}$) 
trained on \datasetname:
\begin{align}
    \mathbf{A_S} = \psi(\mathbf{A_D})
\end{align}

The results (Figs.~\ref{fig:5_method_render}, \ref{fig:5_method_specAlbedo}) 
show how the network differentiates the intensity between hair and skin,
while learning the high-frequency variation that occurs from the pore
occlusion of specular reflection.

\subsection{Specular Normals Extraction}

\paragraph{Background:}
The specular normals exhibit sharp surface details,
such as fine wrinkles and skin pores, 
and are challenging to estimate,
as the appearance of some high-frequency details 
is dependent on the lighting conditions and viewpoint of the texture.
Previous works fail to predict high-frequency details \cite{chen2019photo}, 
or rely on separating the mid- and high-frequency information in two separate maps,
as a generator network may discard the high-frequency as noise \cite{yamaguchi2018high}. 
Instead, we show that it is possible to employ 
an image-to-image translation network with feature matching loss
on a large high-resolution training dataset, 
which produces more detailed and accurate results.

\paragraph{Methodology:}
Similarly to the process for the specular albedo, 
we prefer the diffuse albedo over the reconstructed texture map $\hat{\mathbf{T}}$,
as the latter includes sharp highlights 
that get wrongly interpreted as facial features by the network. 
Moreover, we found that even though the diffuse albedo is stripped from specular reflection, 
it contains the facial skin structures that define mid-~and high-frequency
details, such as pores and wrinkles.
Finally,
since the facial features are similarly distributed across the color channels,
we found that instead of the diffuse albedo $\mathbf{A_D}$,
we can use the luma-transformed (in s\textsc{rgb})
grayscale diffuse albedo ($\mathbf{A_D^{gray}}$).

Again,
we found that the network successfully generates 
both the mid- and high-frequency, when it receives as input
the detailed diffuse albedo $\mathbf{A_D}$
together with the lower-resolution geometry information (in this case, the shape normals).
Moreover, the resulting high-frequency details are more accentuated,
when using normals in tangent space ($\mathbf{N_T}$),
which also serve as a better output,
since most commercial applications require the normals in tangent space.

We train a translation network 
$\rho: \mathbf{A_D^{gray}}, \mathbf{N_T} \mapsto \mathbf{N_S}$,
$\in \mathbb{R}^{512\times512\times3}$ 
to map the concatenation of the grayscale diffuse albedo $\mathbf{A_D^{gray}}$
and the shape normals in tangent space $\mathbf{N_T}$
to the specular normals $\mathbf{N_S}$. 
The specular normals are extracted by the following:
\begin{align}
    \mathbf{N_S} = \rho(\mathbf{A_D^{gray}}, \mathbf{N_T})
\end{align}

\subsection{Diffuse Normals Extraction}
\paragraph{Background:} 
The diffuse normals are highly correlated with the shape normals, 
as diffusion is scattered uniformly across the skin. 
Scars and wrinkles alter the distribution of the diffusion
and some non-skin features such as hair
that do not exhibit significant diffusion.

\paragraph{Methodology :} Similarly to the previous section, 
we train a network
$\sigma: \mathbf{A_D^{gray}}, \mathbf{N_O} \mapsto \mathbf{N_D} 
\in \mathbb{R}^{512\times512\times3}$ 
to map the concatenation of the grayscale diffuse albedo $\mathbf{A_D^{gray}}$ and the shape normals in object space $\mathbf{N_O}$ to the diffuse normals $\mathbf{N_D}$. 
The diffuse normals are extracted as:
\begin{align}
    \mathbf{N_D} = \sigma(\mathbf{A_D^{gray}}, \mathbf{N_O})
\end{align}

Finally, the inferred normals can be used to enhance the reconstructed geometry,
by refining its features and adding plausible details.
We integrate over the specular normals in tangent space 
and produce a displacement map
which can then be embossed on a subdivided base geometry.

\section{Experiments}
\subsection{Implementation Details}

\subsubsection{Patch-Based Image-to-image translation}

\def \width {.19\linewidth}
\begin{figure}[h]
    \centering
    \begin{subfigure}[b]{\width}
        \centering
        \includegraphics[width=\linewidth]{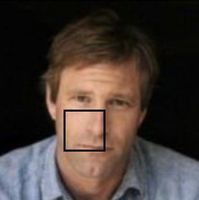}
    \caption{Input}
    \end{subfigure}
    \begin{subfigure}[b]{\width}
        \centering
        \includegraphics[width=\linewidth]{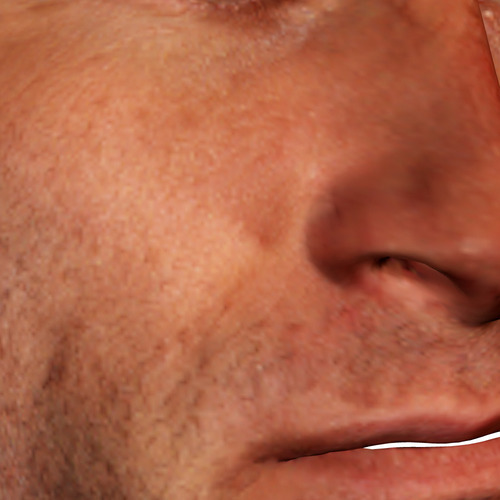}
    \caption{Recon.}
    \end{subfigure}
    \begin{subfigure}[b]{\width}
        \centering
        \includegraphics[width=\linewidth]{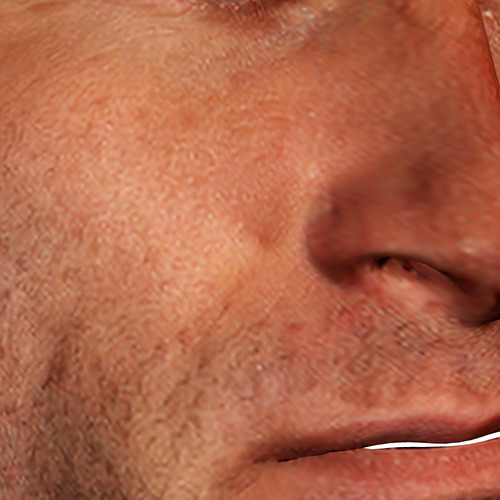}
    \caption{S.R.}
    \end{subfigure}
    \begin{subfigure}[b]{\width}
        \centering
        \includegraphics[width=\linewidth]{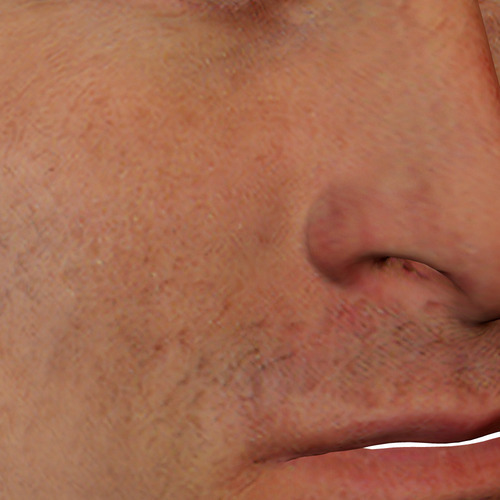}
    \caption{Delight}
    \end{subfigure}
    \begin{subfigure}[b]{\width}
        \centering
        \includegraphics[width=\linewidth]{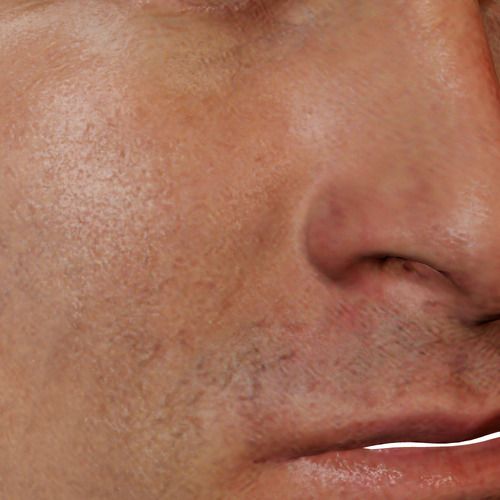}
    \caption{Final}
    \end{subfigure}
    
    \caption{
    Rendering after
    (b) base reconstruction,
    (c) super resolution,
    (d) de-lighting,
    (e) final result.
    }
    \label{fig:process}
\end{figure}

\begin{figure}[]
    \centering
    \begin{subfigure}[b]{.24\linewidth}
        \centering
        \includegraphics[width=\linewidth]{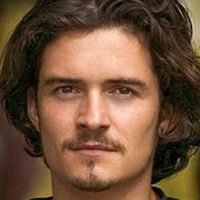}
        \includegraphics[width=\linewidth]{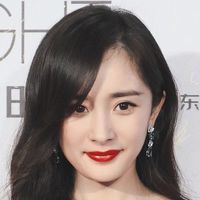}
        \includegraphics[width=\linewidth]{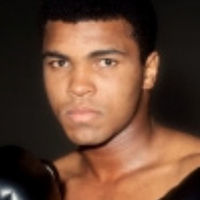}
        \includegraphics[width=\linewidth]{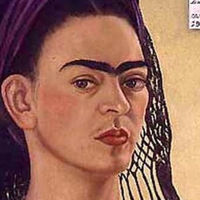}
    \caption{Input}
    \end{subfigure}
    \begin{subfigure}[b]{.24\linewidth}
        \centering
        \includegraphics[width=\linewidth]{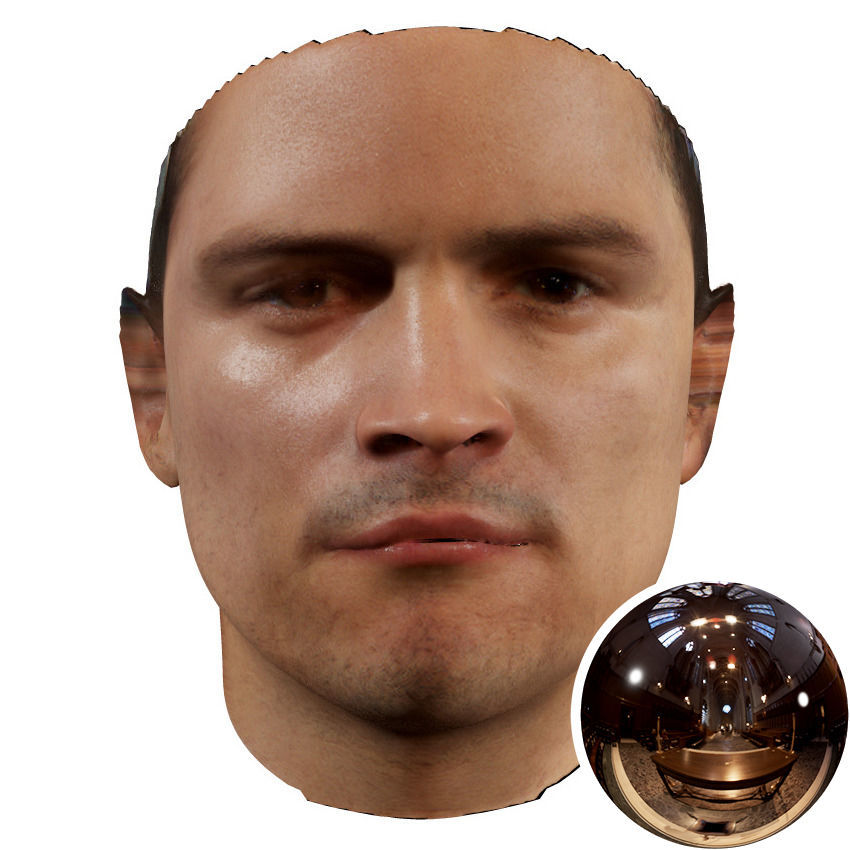}
        \includegraphics[width=\linewidth]{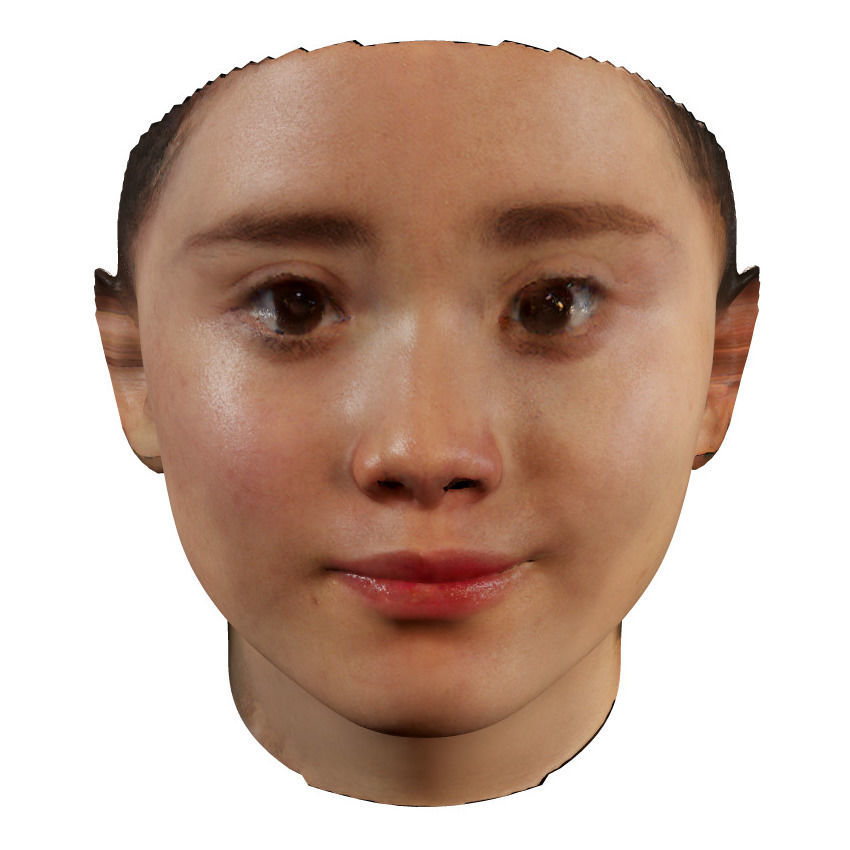}
        \includegraphics[width=\linewidth]{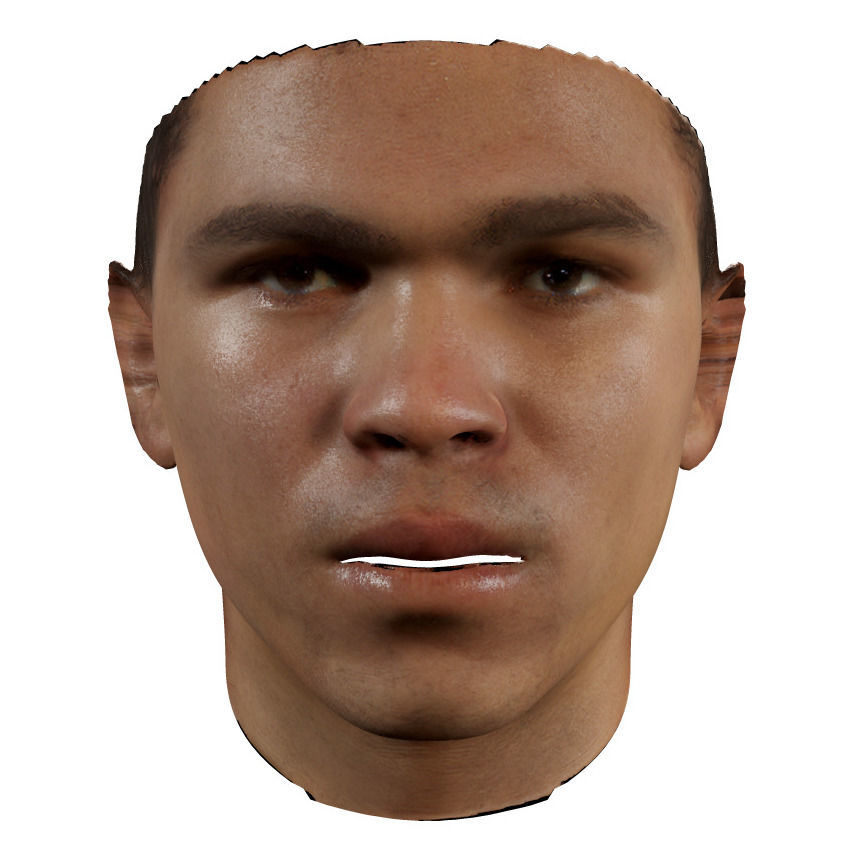}
        \includegraphics[width=\linewidth]{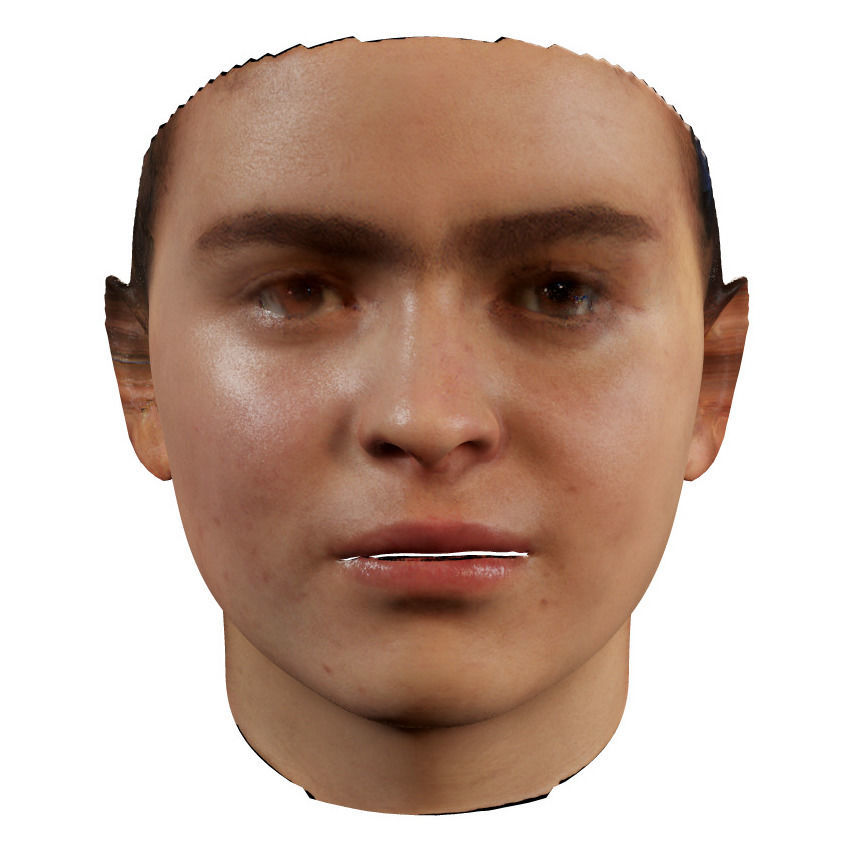}
    \caption{Cathedral}
    \end{subfigure}
    \begin{subfigure}[b]{.24\linewidth}
        \centering
        \includegraphics[width=\linewidth]{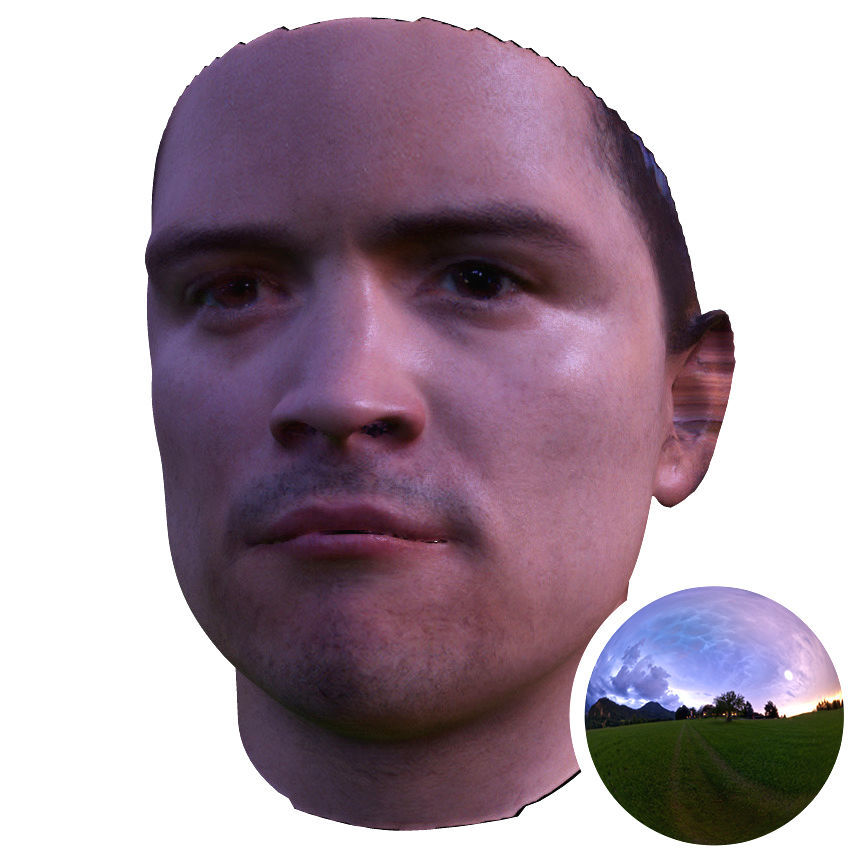}
        \includegraphics[width=\linewidth]{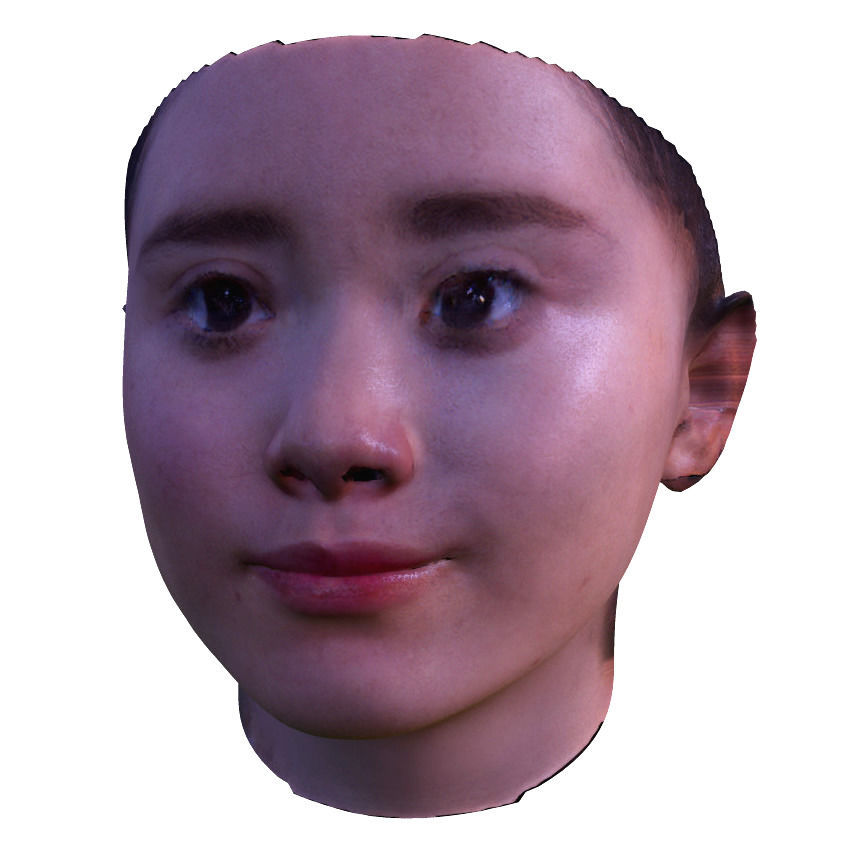}
        \includegraphics[width=\linewidth]{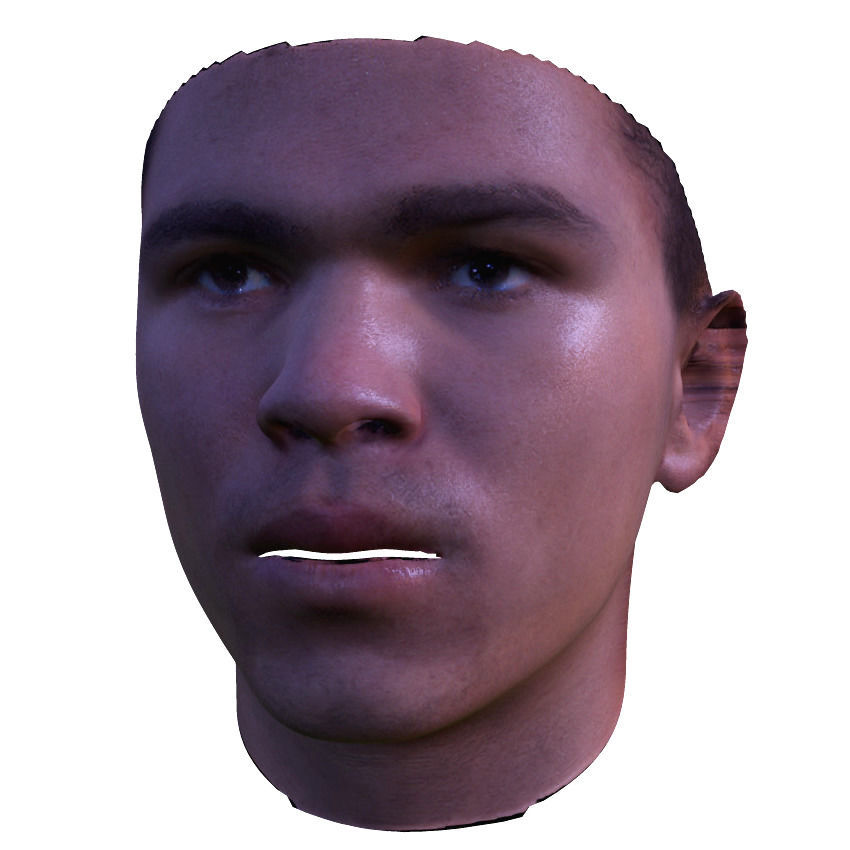}
        \includegraphics[width=\linewidth]{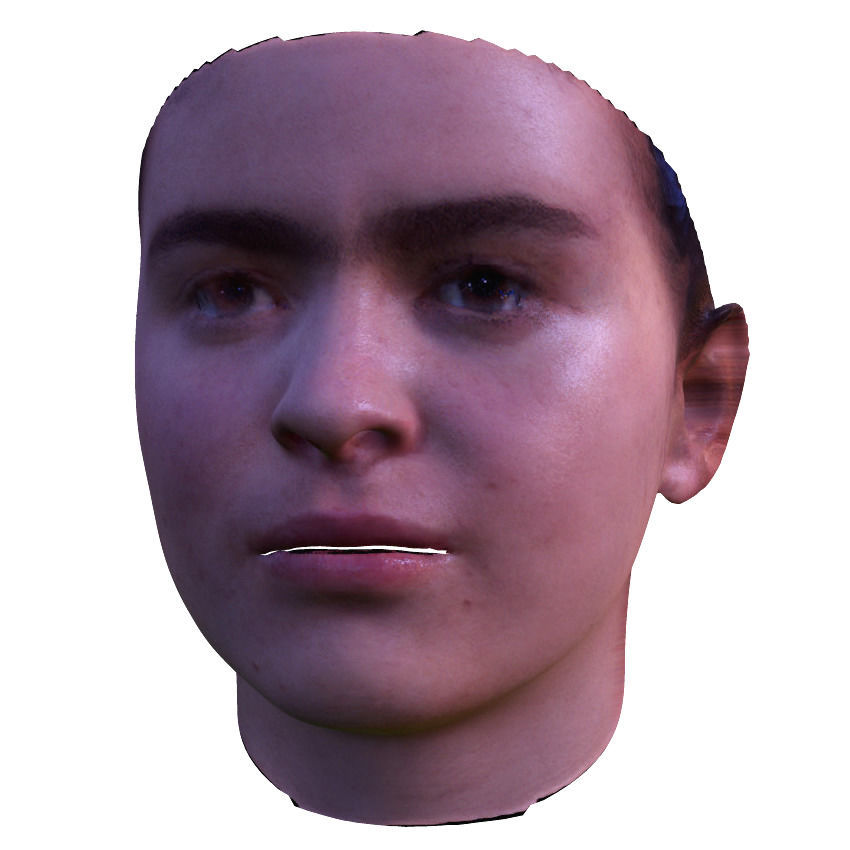}
    \caption{Sunset}
    \end{subfigure}
    \begin{subfigure}[b]{.24\linewidth}
        \centering
        \includegraphics[width=\linewidth]{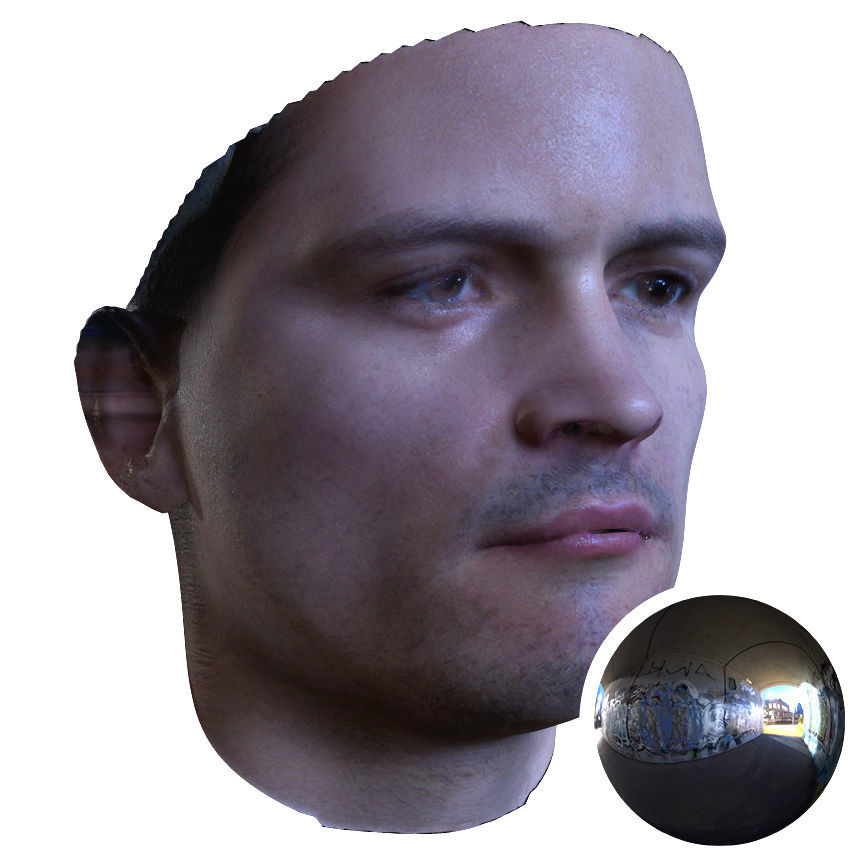}
        \includegraphics[width=\linewidth]{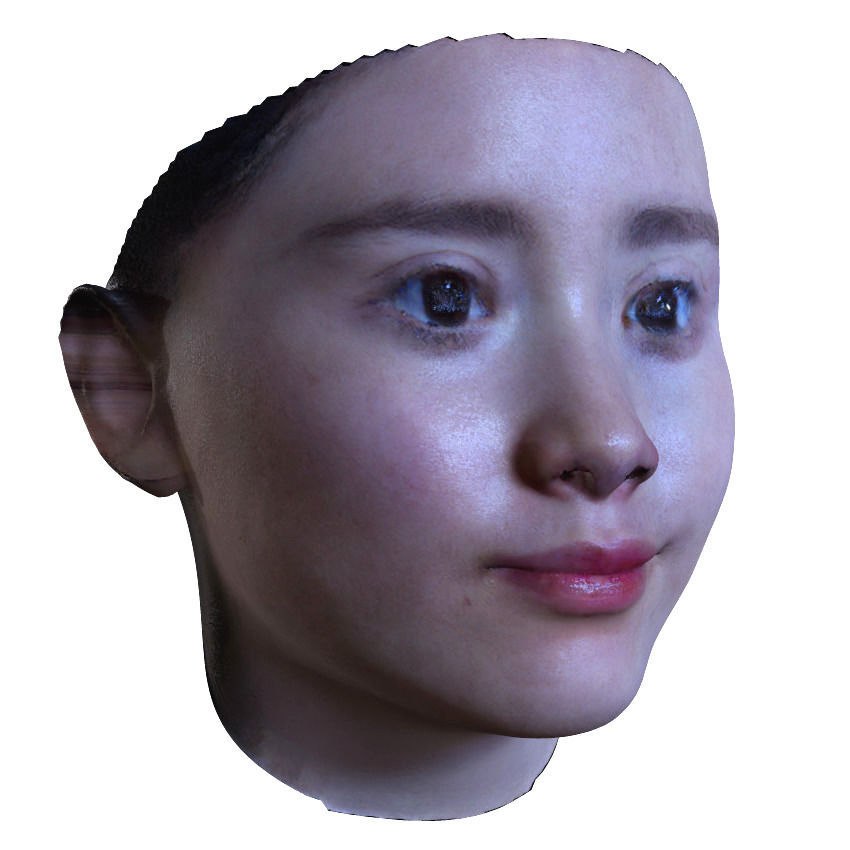}
        \includegraphics[width=\linewidth]{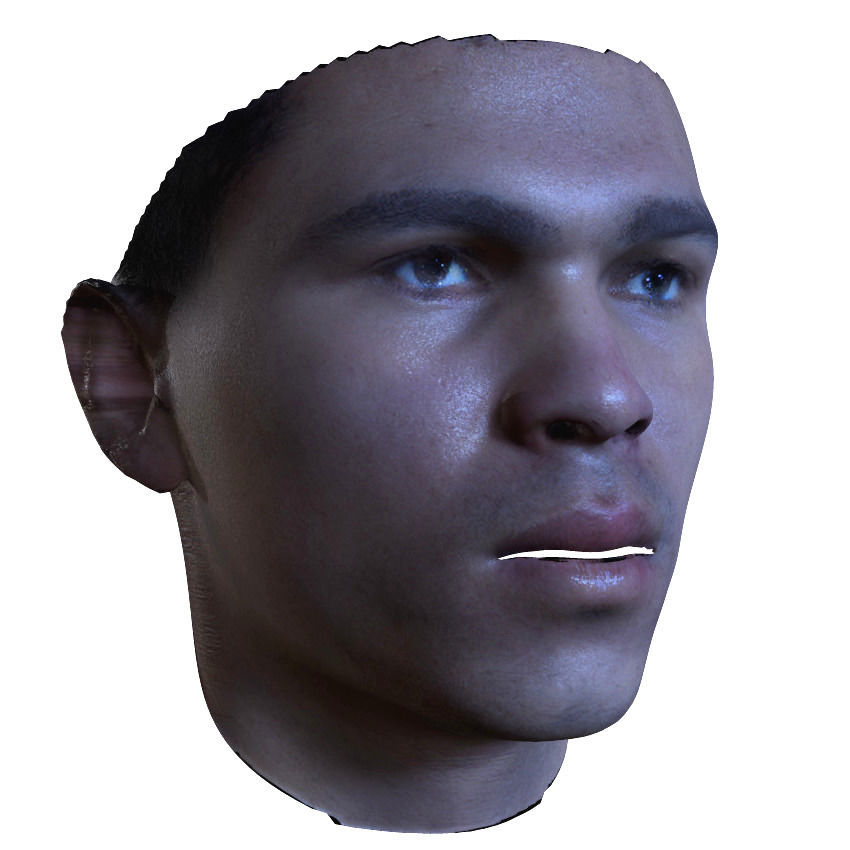}
        \includegraphics[width=\linewidth]{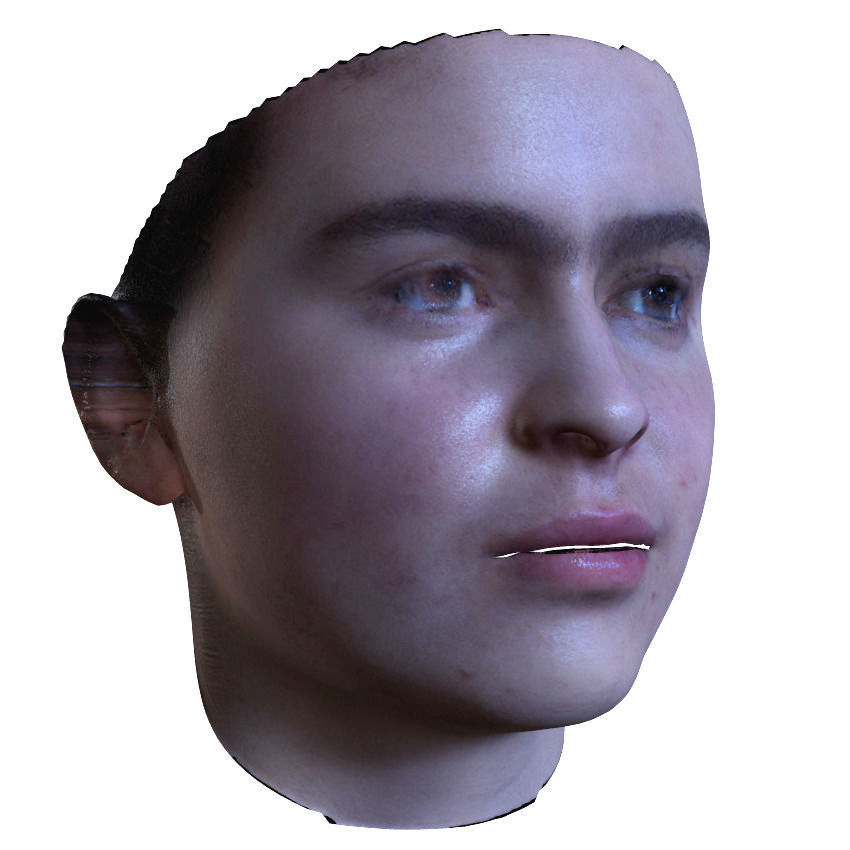}
    \caption{Tunnel}
    \end{subfigure}

    \caption{
        Reconstructions of our method re-illuminated
        under different environment maps \cite{debevec2000acquiring}
        with added spot lights.
    }
    \label{fig:re-illuminated}
\vspace{-0.6cm}
\end{figure}

The tasks of de-lighting,
as well as inferring the diffuse and specular components from a given input image (\textsc{uv})
can be formulated as domain adaptation problems. 
As a result, to carry out the aforementioned tasks 
the model of our choice is pix2pixHD \cite{wang2018high}, 
which has shown impressive results in image-to-image translation
on high-resolution data. 

Nevertheless, as discussed previously: 
(a) our captured data are of very high-resolution (more than \smaller{4}\textsc{k}) 
and thus cannot be used for training ``as-is'' utilizing pix2pixHD,
due to hardware limitations 
(note not even on a \smaller{32}\textsc{gb} \textsc{gpu} we can fit such high-resolution data in their original format), 
(b) pix2pixHD \cite{wang2018high} takes into account only the texture information
and thus geometric details,
in the form of the shape normals and depth cannot be exploited to improve the quality of the generated diffuse and specular components.

To alleviate the aforementioned shortcomings, we: 
(a) split the original high-resolution data into smaller patches of $512\times512$ size. 
More specifically, using a stride of size $256$, 
we derive the partially overlapping patches 
by passing through each original \textsc{uv} horizontally as well as vertically,
(b) for each translation task, 
we utilize the shape normals, 
concatenate them channel-wise with the corresponding grayscale texture input 
(e.g., in the case of translating the diffuse albedo to the specular normals, 
we concatenate the grayscale diffuse albedo with the shape normals channel-wise) 
and thus feed a \fourd{} tensor ($[G,X,Y,Z]$) to the network. 
This increases the level of detail in the derived outputs as the shape normals act as a geometric ``guide''. 
Note that during inference that patch size can be larger (e.g.~$1536\times1536$),
since the network is fully-convolutional.

\vspace{-0.5cm}
\subsubsection{Training Setup}
To train \textsc{rcan} \cite{zhang2018image}, we use the default hyper-parameters.
For the rest of the translation of models, 
we use a custom translation network as described earlier, 
which is based on pix2pixHD \cite{wang2018high}. 
More specifically, we use $9$ and $3$ residual blocks 
in the global and local generators, respectively. 
The learning rate we employed is $0.0001$, 
whereas the Adam betas are $0.5$ for $\beta_1$ and $0.999$ for $\beta_2$. 
Moreover, we do not use the \textsc{vgg} features matching loss 
as this slightly deteriorated the performance. 
Finally, we use as inputs $3$ and $4$ channel tensors 
which include the shape normals $\mathbf{N_O}$ or depth $\mathbf{D_O}$ 
together with the \textsc{rgb} $\mathbf{A_D}$ or grayscale $\mathbf{A_D^{gray}}$ values of the inputs. 
As mentioned earlier, this substantially improves the results by accentuating the details in the translated outputs.

\begin{figure}[t]
    \centering
    \begin{subfigure}[b]{.16\linewidth}
        \centering
        \includegraphics[width=\linewidth]{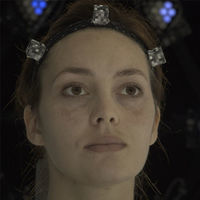}
        \includegraphics[width=\linewidth]{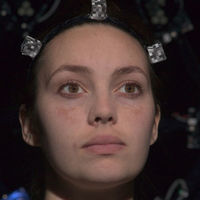}
    \caption{Input}
    \end{subfigure}
    \begin{subfigure}[b]{.24\linewidth}
        \centering
        \includegraphics[width=\linewidth]{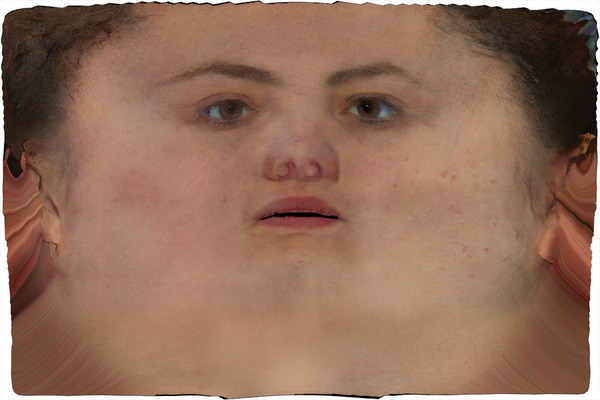}
        \includegraphics[width=\linewidth]{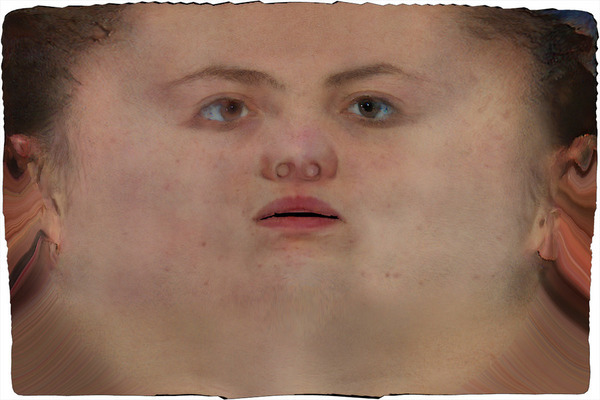}
    \caption{Diff Alb}
    \end{subfigure}
    \begin{subfigure}[b]{.24\linewidth}
        \centering
        \includegraphics[width=\linewidth]{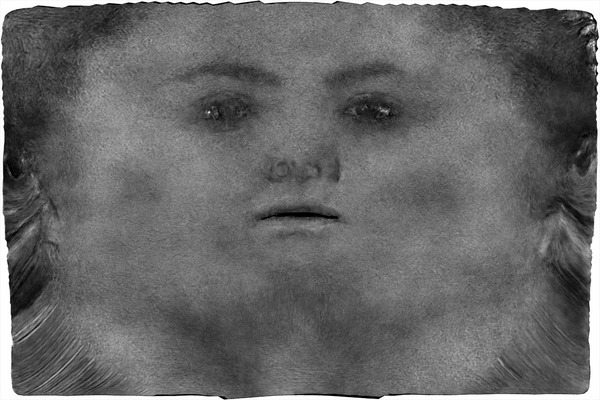}
        \includegraphics[width=\linewidth]{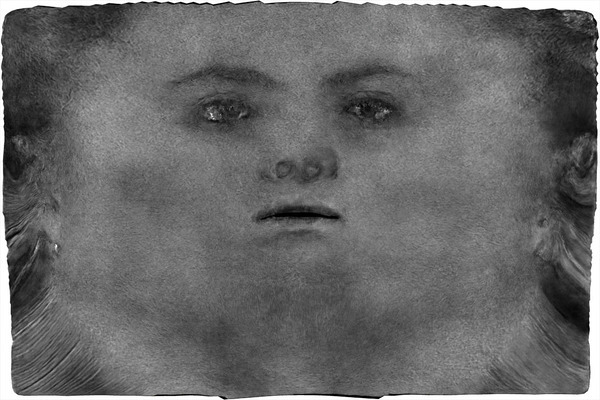}
    \caption{Spec Alb}
    \end{subfigure}
    \begin{subfigure}[b]{.16\linewidth}
        \centering
        \includegraphics[width=\linewidth]{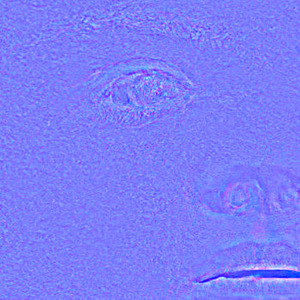}
        \includegraphics[width=\linewidth]{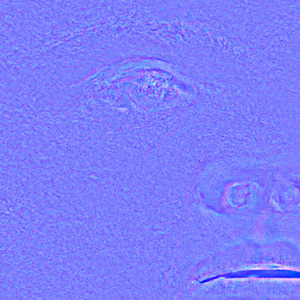}
    \caption{Norm}
    \end{subfigure}
    \begin{subfigure}[b]{.16\linewidth}
        \centering
        \includegraphics[width=\linewidth]{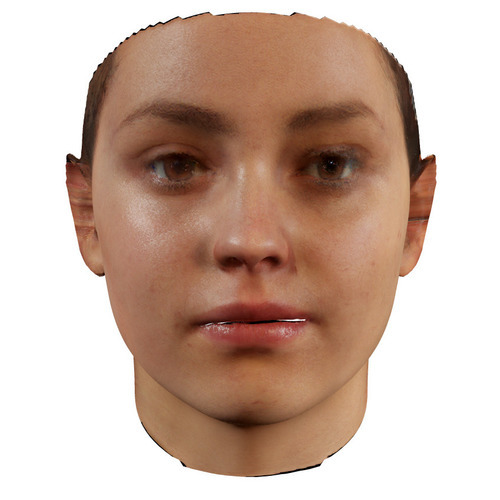}
        \includegraphics[width=\linewidth]{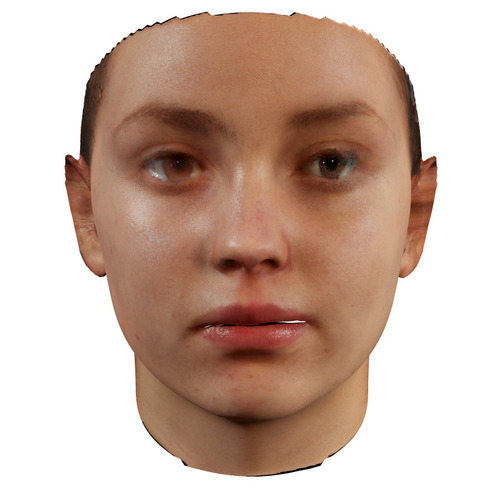}
    \caption{Render}
\end{subfigure}
    \caption{
        Consistency of our algorithm on varying lighting conditions.
        Input images from the Digital Emily Project \cite{alexander2010digital}.
    }
    \label{fig:light_robustness}
\vspace{-0.3cm}
\end{figure}

\def \height {01.7cm}
\def \width {\linewidth}
\begin{figure*}[ht]
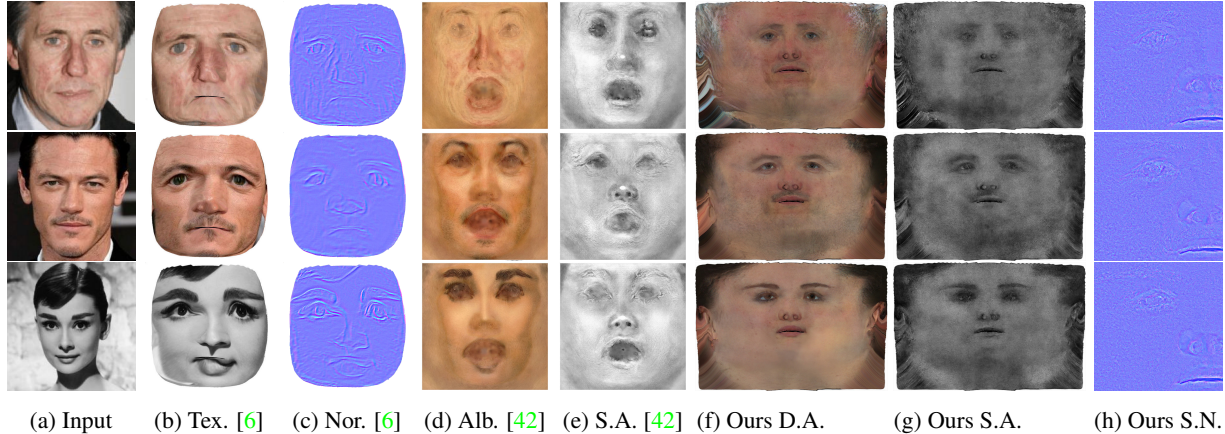

    \centering
    \begin{subfigure}[b]{.1\linewidth}
        \centering
        \foreach \i in {10,7,18}{
        \includegraphics[height=\height, keepaspectratio]{figures/6_results/input/\i.jpg}
        }
    \vspace{-0.4cm}
    \caption{Input}
    \end{subfigure}
    \begin{subfigure}[b]{.1\linewidth}
        \centering
        \foreach \i in {10,7,18}{
        \includegraphics[height=\height, keepaspectratio]{figures/6_results/chen/albedo/\i.jpg}
        }
    \vspace{-0.4cm}
    \caption{Tex. \cite{chen2019photo}}
    \end{subfigure}
    \begin{subfigure}[b]{.1\linewidth}
        \centering
        \foreach \i in {10,7,18}{
        \includegraphics[height=\height, keepaspectratio]{figures/6_results/chen/normals/\i.jpeg}
        }
    \vspace{-0.4cm}
    \caption{Nor. \cite{chen2019photo}}
    \end{subfigure}
    \begin{subfigure}[b]{.1\linewidth}
        \centering
        \foreach \i in {10,7,18}{
        \includegraphics[height=\height, keepaspectratio]{figures/6_results/yamaguchi/diffAlb/\i.jpg}
        }
    \vspace{-0.4cm}
    \caption{Alb. \cite{yamaguchi2018high}}
    \end{subfigure}
    \begin{subfigure}[b]{.1\linewidth}
        \centering
        \foreach \i in {10,7,18}{
        \includegraphics[height=\height, keepaspectratio]{figures/6_results/yamaguchi/specAlb/\i.jpg}
        }
    \vspace{-0.4cm}
    \caption{S.A. \cite{yamaguchi2018high}}
    \end{subfigure}
    \begin{subfigure}[b]{.1\linewidth}
        \centering
        \foreach \i in {10,7,18}{
        \includegraphics[height=\height, keepaspectratio]{figures/6_results/avatarme/diffAlb/\i.jpg}
        }
    \vspace{-0.4cm}
    \caption{Ours D.A.}
    \end{subfigure}
    \hspace{0.72cm}
    \begin{subfigure}[b]{.1\linewidth}
    \captionsetup{justification=raggedright,singlelinecheck=false}
        \centering
        \foreach \i in {10,7,18}{
        \includegraphics[height=\height, keepaspectratio]{figures/6_results/avatarme/specAlb/\i.jpg}
        }
    \vspace{-0.4cm}
    \caption{Ours S.A.}
    \end{subfigure}
    \hspace{0.72cm}
    \begin{subfigure}[b]{.1\linewidth}
        \centering
        \foreach \i in {10,7,18}{
        \includegraphics[height=\height, keepaspectratio]{figures/6_results/avatarme/specNorms/\i.jpg}
        }
    \vspace{-0.4cm}
    \caption{Ours S.N.}
\end{subfigure}
\caption{
    Comparison of reflectance maps predicted by our method against state-of-the-art methods.
    \cite{yamaguchi2018high} reconstruction \\is provided by the authors and \cite{chen2019photo} from their open-sourced models. Last column is cropped to better show the details.
} 
\label{fig:comparisons}
\end{figure*}

\subsection{Evaluation}
\begin{figure}[h!]
    \centering
    \begin{subfigure}[b]{.24\linewidth}
        \centering
        \includegraphics[height=2cm]{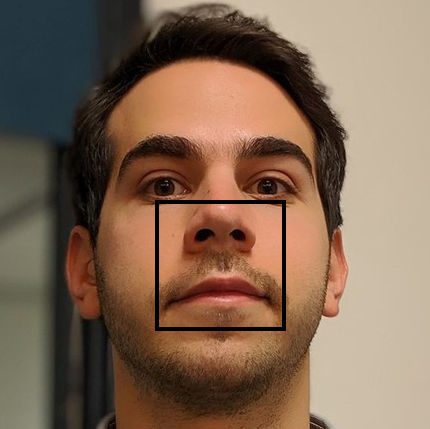}
        \includegraphics[height=2cm]{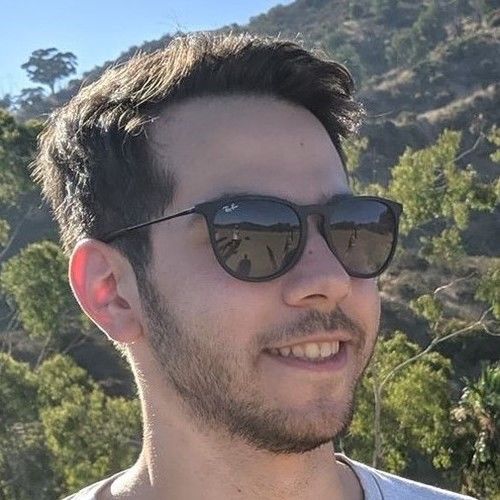}
    \caption{Input}
    \end{subfigure}
    \begin{subfigure}[b]{.24\linewidth}
        \centering
        \includegraphics[height=2cm]{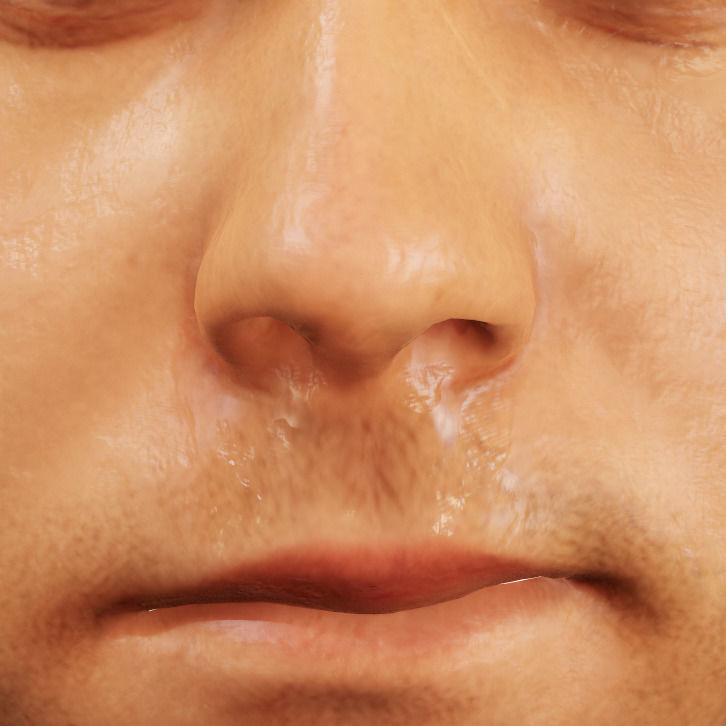}
        \includegraphics[height=2cm]{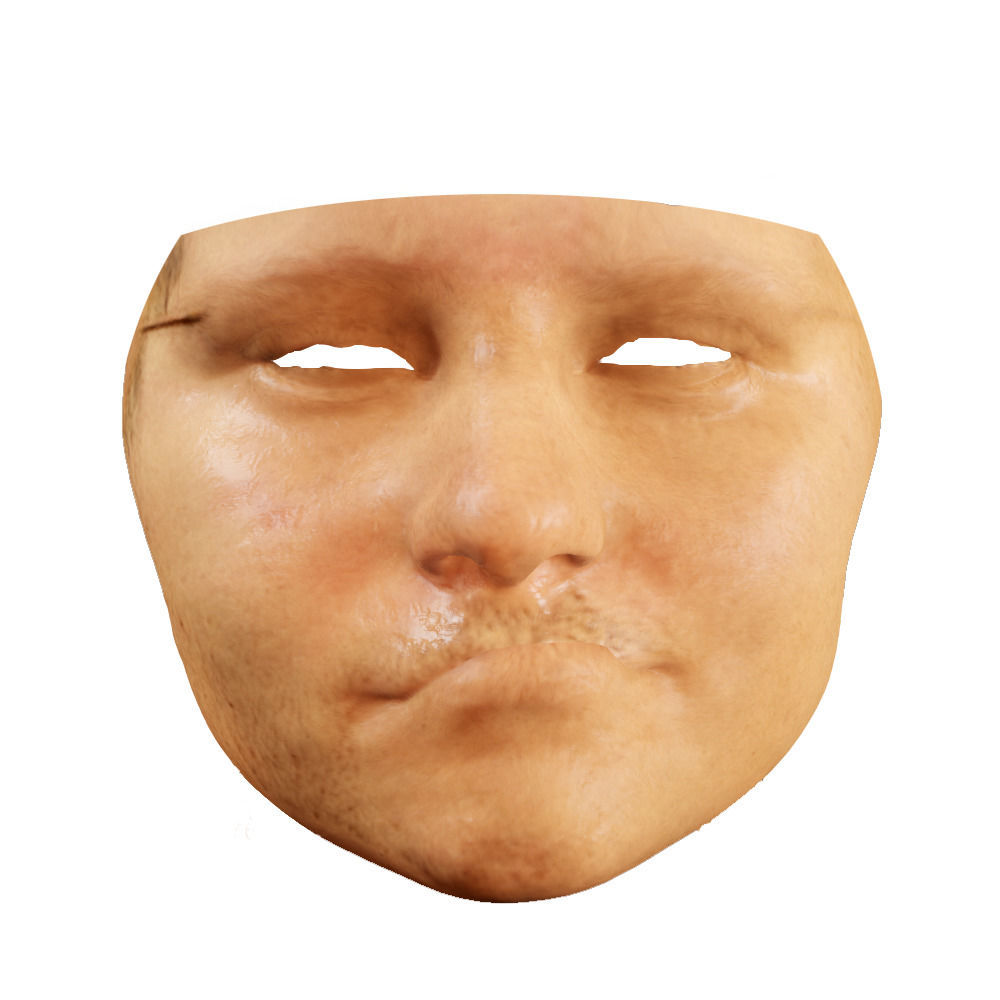}
    \caption{\cite{yamaguchi2018high}}
    \end{subfigure}
    \begin{subfigure}[b]{.24\linewidth}
        \centering
        \includegraphics[height=2cm]{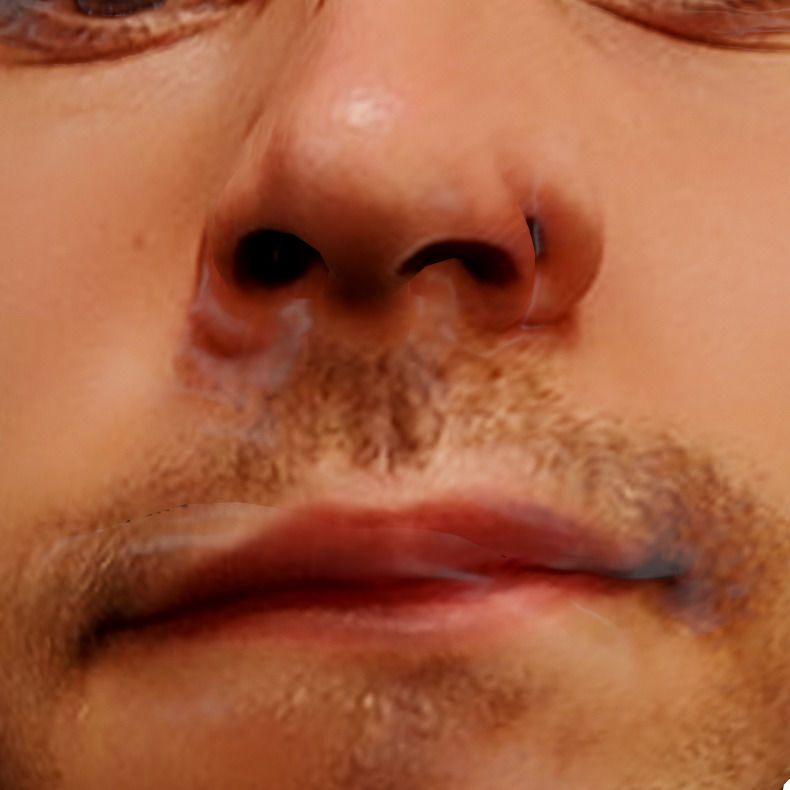}
        \includegraphics[height=2cm]{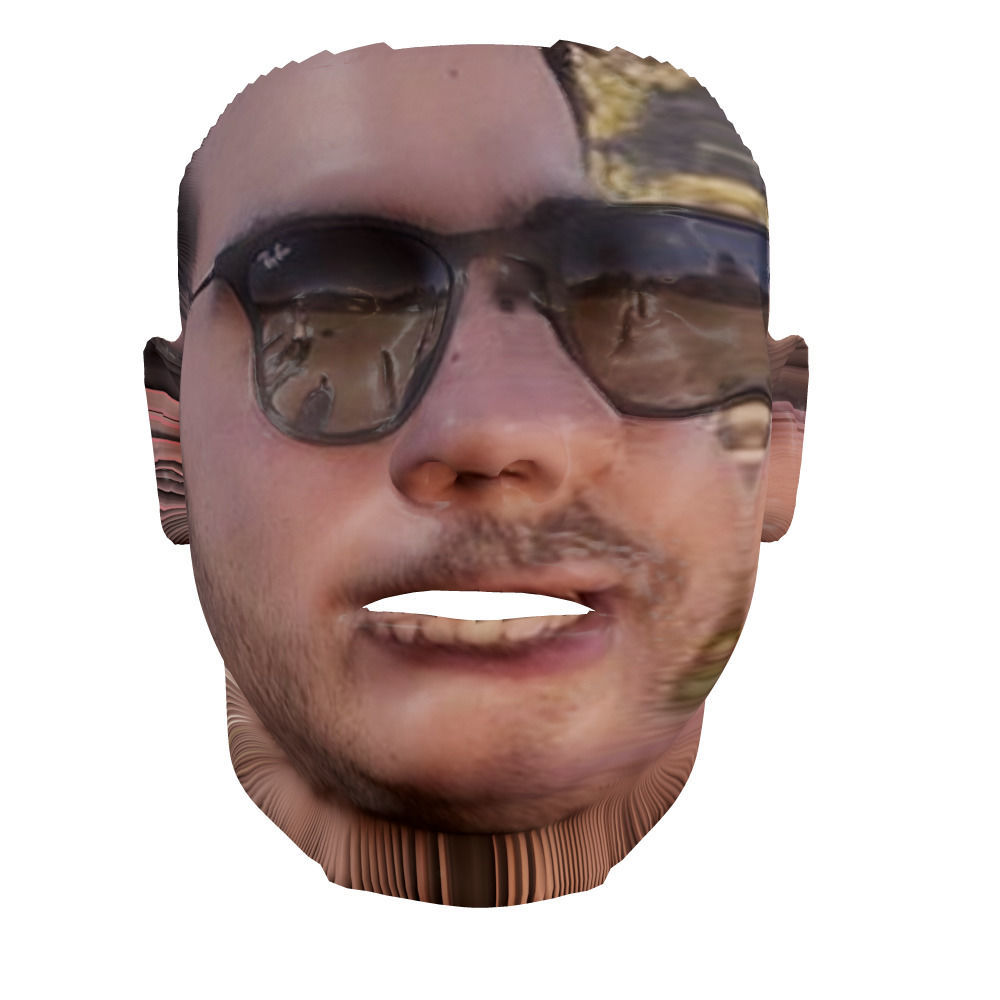}
    \caption{\cite{chen2019photo}}
    \end{subfigure}
    \begin{subfigure}[b]{.24\linewidth}
        \centering
        \includegraphics[height=2cm]{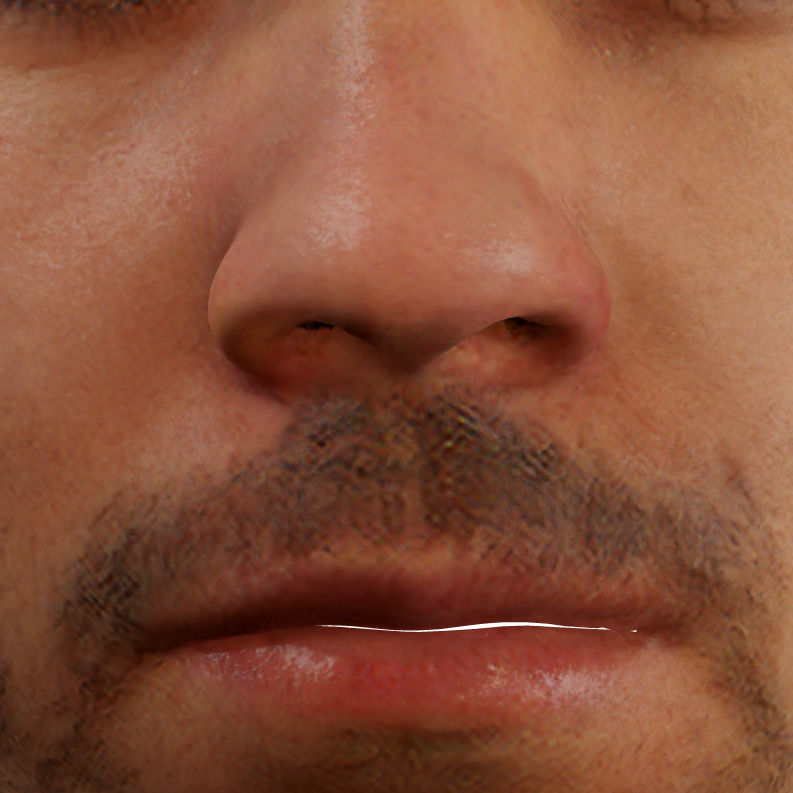}
        \includegraphics[height=2cm]{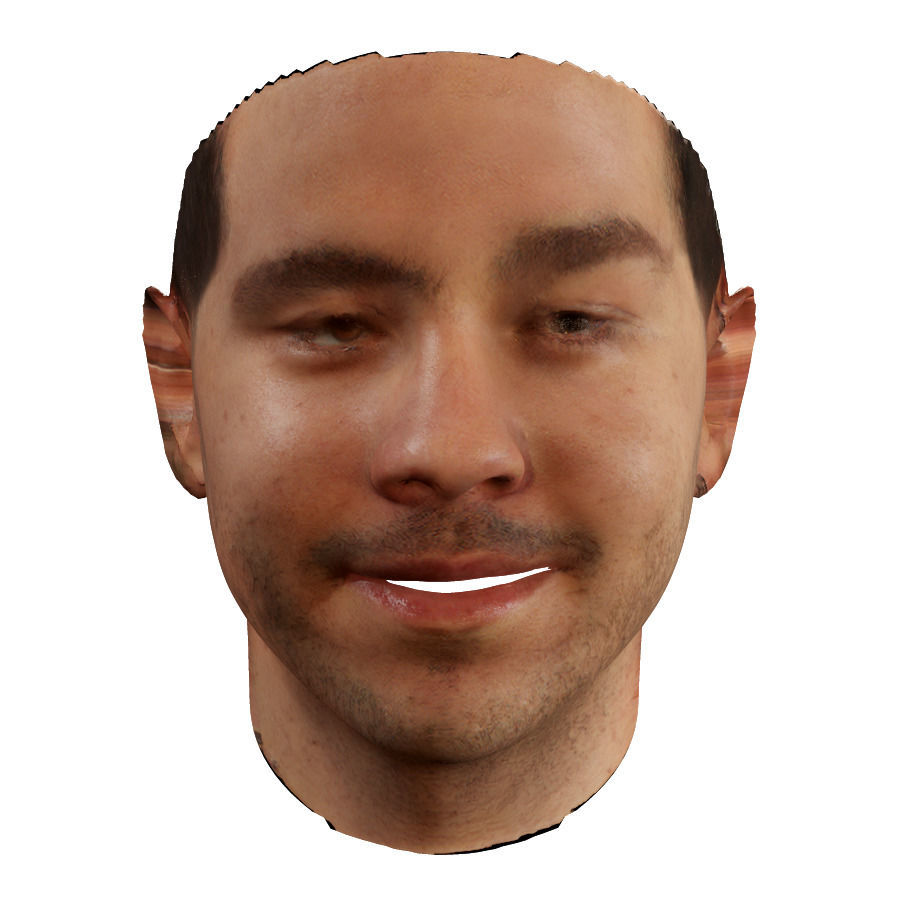}
    \caption{Ours}
    \end{subfigure}
    \caption{
        Qualitative comparison of reconstructions of a subject from ``in-the-wild`` images, rendered in the Grace Cathedral environment \cite{debevec2000acquiring}.
        \cite{yamaguchi2018high} reconstructions provided by the authors and \cite{chen2019photo} from their open-sourced models.
    }
    \label{fig:light_robustness_2}
\begin{minipage}[b]{\linewidth}
\vspace{0.6cm}
\begin{tabular}{|l|lll|}
\hline
Algorithm       & \cite{yamaguchi2018high}  & \cite{chen2019photo} & Ours           \\ 
\hline
\textsc{psnr} (Albedo)    & 11.225              & 14.374        & \textbf{24.05}  \\
\textsc{psnr} (Normals)   & 21.889              & 17.321        & \textbf{26.97} \\
Rendered ID Match \cite{deng2019arcface} &  0.632  & 0.629         & \textbf{0.873} \\
\hline

\end{tabular}
\captionof{table}{
    Average \textsc{psnr} computed for a single subject between 6 reconstructions
    of the same subject from ``in-the-wild`` images and the ground truth captures with \cite{lattas2019multi}.
    We transform \cite{chen2019photo, yamaguchi2018high} results to our \textsc{uv} topology and compute only for a $2K\times2K$ centered crop,
    as they only produced the frontal part of the face
    and manually add eyes to \cite{yamaguchi2018high}.
}\label{table:comparisons}
\end{minipage}
\end{figure}

We conduct quantitative as well as qualitative comparisons against the state-of-the-art. 
For the quantitative comparisons, 
we utilize the widely used \textsc{psnr} metric \cite{hore2010image}, 
and report the results in Table \ref{table:comparisons}. 
As can be seen, our method outperforms \cite{chen2019photo} and \cite{yamaguchi2018high} by a significant margin. 
Moreover using a state-of-the-art face recognition algorithm \cite{deng2019arcface},
we also find the highest match of facial identity compared to the input images when using our method. 
The input images were compared against renderings of the faces 
with reconstructed geometry and reflectance, including eyes.

For the qualitative comparisons, 
we perform \threed{} reconstructions of ``in-the-wild'' images. 
As shown in Figs.~\ref{fig:comparisons} and \ref{fig:light_robustness_2},
our method does not produce any artifacts in the final renderings 
and successfully handles extreme poses and occlusions such as sunglasses. 
We infer the texture maps in a patch-based manner from high-resolution input, 
which produces higher-quality details than \cite{chen2019photo, yamaguchi2018high}, 
who train on high-quality scans but infer the maps for the whole face, 
in lower resolution. 
This is also apparent in Fig.~\ref{fig:process}, 
which shows our reconstruction after each step of our process.
Moreover, we can successfully acquire each component from
black-and-white images (Fig.~\ref{fig:light_robustness_2})
and even drawn portraits (Fig.~\ref{fig:comparisons}).

Furthermore, we experiment with different environment conditions,
in the input images and while rendering.
As presented in Fig.~\ref{fig:light_robustness}, 
the extracted normals, diffuse and specular albedos are consistent, 
regardless of the illumination on the original input images. 
Finally, Fig.~\ref{fig:re-illuminated} 
shows different subjects rendered under different environments.
We can realistically illuminate each subject in each scene and
accurately reconstruct the environment reflectance,
including detailed specular reflections and subsurface scattering.

In addition to the facial mesh,
we are able to infer the entire head topology
based on the Universal Head Model (\textsc{uhm}) \cite{ploumpis2019towards,ploumpis2019combining}.
We project our facial mesh to a subspace, regress the head latent parameters and then finally derive the completed head model with completed textures.
Some qualitative head completion results can be seen in Figs~\ref{fig:teaser}, \ref{fig:overview}.

\subsection{Limitations} 
While our dataset contains a relatively large number of subjects,
it does not contain sufficient examples of subjects from certain ethnicities.
Hence, our method currently does not perform that well when we reconstruct faces of e.g.~darker skin subjects.
Also, the reconstructed specular albedo and normals exhibit 
slight blurring of some high frequency pore details 
due to minor alignment errors of the acquired data to the template \threedmm{} model. 
Finally, the accuracy of facial reconstruction is not completely independent of the quality of the input photograph, 
and well-lit, higher resolution photographs produce more accurate results.
\section{Conclusion}
In this paper, 
we propose the first methodology that produces high-quality 
rendering-ready face reconstructions from arbitrary ``in-the-wild'' images. 
We build upon recently proposed \threed{} face reconstruction techniques 
and train image translation networks that can perform estimation of high quality 
(a) diffuse and specular albedo, and 
(b) diffuse and specular normals. 
This is made possible with a large training dataset of 200 faces 
acquired with high quality facial capture techniques. 
We demonstrate that it is possible to produce rendering-ready faces from arbitrary face images varying in pose, occlusions, etc., 
including black-and-white and drawn portraits. 
Our results exhibit unprecedented level of detail and realism in the reconstructions,
while preserving the identity of subjects in the input photographs.

\FloatBarrier


\section*{Acknowledgements}
AL was supported by EPSRC Project DEFORM (EP/S010203/1) and
SM by an Imperial College FATA.
AG acknowledges funding by the EPSRC Early Career Fellowship (EP/N006259/1)
and SZ from a Google Faculty Fellowship 
and the EPSRC Fellowship DEFORM (EP/S010203/1).

{\small
\bibliographystyle{ieee_fullname}
\bibliography{egbib}
}

\end{document}